\documentclass{article}

\usepackage{microtype}
\usepackage{graphicx}
\usepackage{subcaption}
\usepackage{booktabs} % for professional tables

\usepackage{hyperref}

\usepackage[preprint]{icml2026}

\usepackage{amsmath}
\usepackage{amssymb}
\usepackage{mathtools}
\usepackage{amsthm}

\usepackage{url}

\usepackage[utf8]{inputenc} % allow utf-8 input
\usepackage[T1]{fontenc}

\usepackage{booktabs}       % professional-quality tables
\usepackage{amsfonts}       % blackboard math symbols
\usepackage{nicefrac}       % compact symbols for 1/2, etc.
\usepackage{microtype}      % microtypography
\usepackage[dvipsnames]{xcolor}         % colors

\usepackage{amsmath}
\usepackage{multirow}
\usepackage{pifont}
\usepackage{listings}
\usepackage{caption}
\usepackage{bbm}
\usepackage{arydshln}
\usepackage{listings}
\usepackage{placeins}
\usepackage{nicefrac}
\usepackage{xurl}
\usepackage{enumitem}
\usepackage{tabularx}
\usepackage{array}
\usepackage{nicefrac}
\usepackage{makecell}
\usepackage{xspace}
\usepackage{graphbox}
\usepackage{pifont}
\usepackage{enumitem}
\usepackage{colortbl}
\usepackage{multirow}
\usepackage{diagbox}
\usepackage{array}
\usepackage{wrapfig}
\usepackage{subcaption}
\usepackage{footmisc}
\usepackage{color}
\usepackage{stfloats}
\usepackage{slashbox}
\usepackage{diagbox}
\usepackage{float}
\usepackage{mathrsfs}
\usepackage{apptools}
\usepackage{bbm}
\usepackage{soul}
\usepackage[export]{adjustbox}
\usepackage{fontawesome5}

\usepackage[capitalize,noabbrev]{cleveref}

\theoremstyle{plain}

\theoremstyle{definition}

\theoremstyle{remark}

\newcommand{\norm}[1]{\left\|#1\right\|}

\definecolor{brightpurple}{RGB}{160,0,255}
\newcommand{\rev}[1]{%\textcolor{brightpurple}{#1}
#1
\xspace}

\newcommand{\myparagraph}[1]{\noindent\textbf{#1}}

\newcolumntype{C}[1]{>{\centering\arraybackslash}p{#1}}
\newcolumntype{L}[1]{>{\raggedright\arraybackslash}p{#1}}
\newcolumntype{R}[1]{>{\raggedleft\arraybackslash}p{#1}}
\newlength\newl
\newlength\newlc

\newlength\colwidth
\newlength\figwidth

\usepackage{amsmath,amsfonts,bm}

\def\eqref#1{(\ref{#1})}
\def\1{\bm{1}}

\DeclareMathAlphabet{\mathsfit}{\encodingdefault}{\sfdefault}{m}{sl}
\SetMathAlphabet{\mathsfit}{bold}{\encodingdefault}{\sfdefault}{bx}{n}

\DeclareMathOperator*{\argmin}{arg\,min}

\newcommand{\imnet}{ImageNet\xspace}

\newcommand{\titok}{TiTok\xspace}
\newcommand{\flextok}{FlexTok\xspace}
\newcommand{\unitok}{UniTok\xspace}

\newcommand{\fuselip}{FuseLIP\xspace}
\newcommand{\fuselipS}{\fuselip-S\xspace}
\newcommand{\fuselipB}{\fuselip-B\xspace}

\newcommand{\cciii}{CC3M\xspace}

\usepackage[textsize=tiny]{todonotes}

\icmltitlerunning{%Submission and Formatting Instructions for ICML 2026
On the Adversarial Robustness of Discrete Image Tokenizers}

\begin{document}

\twocolumn[
  \icmltitle{On the Adversarial Robustness of Discrete Image Tokenizers}



  \icmlsetsymbol{equal}{*}

  \begin{icmlauthorlist}
    \icmlauthor{Rishika Bhagwatkar}{yyy}
    \icmlauthor{Irina Rish}{yyy}
    \icmlauthor{Nicolas Flammarion}{comp}
    \icmlauthor{Francesco Croce}{aalto}
  \end{icmlauthorlist}

  \icmlaffiliation{yyy}{Mila - Quebec AI Institute}
  \icmlaffiliation{comp}{EPFL}
  \icmlaffiliation{aalto}{ELLIS Institute Finland - Aalto University}

  \icmlcorrespondingauthor{Rishika Bhagwatkar}{rishika.bhagwatkar@mila.quebec}

  \icmlkeywords{Machine Learning, ICML}

  \vskip 0.3in
]



\printAffiliationsAndNotice{}  % no special notice (required even if empty)

\begin{abstract} 
  Discrete image tokenizers encode visual inputs as sequences of tokens from a finite vocabulary and are gaining popularity in multimodal systems, including encoder-only, encoder-decoder, and decoder-only models. However, unlike CLIP encoders, their vulnerability to adversarial attacks has not been explored. Ours being the first work studying this topic, we first formulate attacks that aim to perturb the features extracted by discrete tokenizers, and thus change the extracted tokens. These attacks are computationally efficient, application-agnostic, and effective across classification, multimodal retrieval, and captioning tasks. Second, to defend against this vulnerability, inspired by recent work on robust CLIP encoders, we fine-tune popular tokenizers with unsupervised adversarial training,  keeping all other components frozen. While unsupervised and task-agnostic, our approach significantly improves robustness to both unsupervised and end-to-end supervised attacks and generalizes well to unseen tasks and data. Unlike supervised adversarial training, our approach can leverage unlabeled images, making it more versatile. Overall, our work highlights the critical role of tokenizer robustness in downstream tasks and presents an important step in the development of safe multimodal foundation models.
\href{https://robust-tokenizers.github.io/}{
    \faGlobe\ \texttt{robust-tokenizers.github.io}
}

  % Project webpage: \href{https://robust-tokenizers.github.io/}{https://robust-tokenizers.github.io}.\looseness-1
\end{abstract}

\section{Introduction}

Similar to text tokenizers, discrete image tokenizers represent visual input as a sequence, typically of fixed length, of vectors from a finite codebook \cite{vandenoord2017neural, yu2024titok, ma2025unitok, bachmann2025flextok}.
However, unlike text tokenizers such as BPE \cite{bpe}, they rely on deep networks trained alongside a de-tokenization model for image reconstruction.
% they rely on deep networks, which are trained, together with a de-tokenization model, for image reconstruction.
These de-tokenizers enable synthetic image generation in autoregressive frameworks \cite{ma2025unitok, xie2025showo, wang2024emu3}, which are competitive with diffusion models.
Recently, discrete image tokenizers have become integral to complex architectures as image encoders, offering a popular alternative to CLIP \cite{radford2021clip} and DINO \cite{oquab2023dinov2}.
Multimodal encoder-decoder models such as Unified-IO \cite{lu2022unified}, 4M \cite{mizrahi20234m} and 4M-21 \cite{bachmann20244m21} employ discrete tokenizers for natural images and segmentation maps, while \fuselip \cite{schlarmann2025fuselip} combines them with early-fusion in text-image models trained with contrastive objectives.
% leverages them to enable a  multimodal embedding model, based on early-fusion of text and image inputs, trained with a contrastive loss.
% These models can be applied, either zero-shot or via transfer learning, on a variety of unimodal and multimodal tasks, including classification, retrieval, and VQA.
These models can be applied to various unimodal and multimodal tasks, including classification, retrieval, and VQA.
% Moreover, discrete image tokenizers are used by decoder-only generative models \cite{team2024chameleon, xie2025showo, wang2025selftok, ma2025unitok}.
% In fact, the finite vocabulary enables language-style autoregressive modeling, yielding unified understanding and generation in both vision and language domains.
Additionally, decoder-only generative models \cite{team2024chameleon, xie2025showo, wang2025selftok, ma2025unitok} leverage discrete tokenizers for unified vision and language understanding and generation. %in both domains 
%
% As a consequence, the robustness (or lack of it) of image tokenizers to adversarial attacks directly influences how vulnerable all such models are.
Consequently, the robustness of image tokenizers to adversarial attacks directly impacts the vulnerability of all models using them.
% However, while several works have focused on testing and improving the adversarial robustness of standard, (e.g.,~CLIP), encoders, this aspect of discrete image tokenizers remain completely unexplored.
While much research has focused on improving the adversarial robustness of standard encoders (e.g., CLIP), this aspect of discrete image tokenizers remains completely unexplored.

In this work, we first study the vulnerability of recently proposed discrete image tokenizers, both in isolation and as part of larger models.
In particular, we test \textit{unsupervised attacks} which aim to alter the tokenization of the original image by distorting the encoding and consequently assigning incorrect codebook vectors.
Since these attacks target only the discrete image tokenizer, the resulting perturbations are agnostic to the downstream tasks for which the tokens are used (e.g., reconstruction, classification, language generation).
We show that such direct unsupervised attacks are often effective against standard tokenizers, and in some cases even competitive with end-to-end supervised attacks which require task-specific information (e.g., labels) and are more computationally expensive because they target the entire system.
For instance, in autoregressive multimodal models, the large language model (LLM) has one to two orders of magnitude more parameters than the image tokenizer.
Notably, our unsupervised attacks can make an LLM output a target (malicious) caption for a given input image by matching its tokenization %of the input image 
to that of a target image semantically aligned with the desired caption, without requiring direct access to the LLM itself.
These results highlight the vulnerability of existing discrete image tokenizers, and the safety risks they pose to downstream models across %a variety of 
tasks. %\looseness-1

Then, in order to mitigate this vulnerability, we extend the framework of \citet{schlarmann2024robust}, introduced for robust CLIP vision encoders, to discrete image tokenizers.
In particular, we fine-tune discrete image tokenizers with unsupervised adversarial training, i.e.,~we train the model to yield consistent tokenization for the original images and their corresponding adversarial counterparts, computed on-the-fly by unsupervised attacks.
Since this approach is independent of any downstream task, the resulting tokenizer can be directly plugged back into any system that relied on the original tokenizer without the need for additional adaptations.
In fact, through extensive experiments, we show that unsupervised adversarial fine-tuning increases the robustness to end-to-end supervised attacks in a variety of tasks, from image classification to captioning.
By fine-tuning \titok tokenizers, we obtain robust \fuselip models for multimodal embedding, while with adversarial training of \unitok we get a multimodal LLM (MLLM) robust on VQA and captioning tasks.
Finally, we provide fine-grained %comparisons to 
analyses of the effect of unsupervised attacks by reconstructing the adversarial images, which reveal quite significant differences across tokenizers, %the differences between
supervised vs.\ unsupervised adversarial training, and the role of the dataset used for adversarial fine-tuning.\looseness-1

\myparagraph{Contributions.} In summary, (a) our work is the first to systematically test and improve the adversarial robustness of discrete image tokenizers. (b) To evaluate their vulnerability, we propose unsupervised attacks that are both efficient and task-agnostic. (c) Importantly, we also show that the same attacks can be leveraged to adversarially fine-tune these tokenizers. (d) Our unsupervised adversarial fine-tuning robustifies the tokenizers against both unsupervised and end-to-end supervised attacks, at a significantly lower computational cost than end-to-end supervised adversarial fine-tuning. (e) In contrast to supervised defenses, our defense can directly leverage any amount of unlabeled data. (f) These robust tokenizers can be seamlessly integrated as image encoders in larger systems, strengthening the robustness of multimodal embedding models (\fuselip) and MLLMs (\unitok-MLLM) to all attacks (supervised or not) across diverse tasks well beyond the training data. %Therefore, our work identifies the crucial role of tokenizers in safeguarding multimodal models while offering a practical step toward building more robust and safe foundation models.

% This yields, at a low computational cost, robust tokenizers which can be deployed in larger systems using them as image encoders.

% In this way we could improve, for example, the resistance to end-to-end attacks of multimodal embedding models (\fuselip) as well as LLMs (\unitok-MLLM) on a variety of tasks.
% Therefore, our approach constitutes a practical step towards more robust and safe foundation models.

%These discrete tokens support several downstream applications, including generative modeling, reconstruction, classification and retrieval. For reconstruction, a decoder is provided with the codes to reconstruct the image. In generative models, the finite vocabulary enables language-style autoregressive or masked modeling over token indices. Representations for classification and retrieval tasks can be obtained by pooling the codes. Owing to their several applications, discrete image tokenizers are becoming an increasingly popular alternative to CLIP image encoders for multimodal models, including encoder-only (FuseLIP), encoder-decoder (4M, 4M-21) and decoder-only (Chameleon, Show-o) models. Yet there are very few studies investigating the adversarial robustness of these discrete tokenizers.

% Preamble:
% \usepackage{algorithm}
% \usepackage{algpseudocode}
% \usepackage{amsmath}

\section{Related Work}

\noindent\textbf{Image tokenizers.}
Image tokenizers aim to compress visual data via deep learning models, trained with image reconstruction objectives possibly aided by auxiliary losses.
Continuous tokenizers encode images into a continuous feature space \cite{fan2025unified}.
Conversely, discrete tokenizers convert images into sequences of tokens drawn from a fixed vocabulary, typically via vector quantization.
VQ-VAE \cite{vandenoord2017neural} introduced a vector-quantized bottleneck, replacing continuous latents with discrete codes,
%Subsequent improvements included adversarial training (VQGAN \cite{yu2022vectorquantized}), transformer architectures (ViT-VQGAN \cite{yu2022vectorquantized}, Efficient-VQGAN \cite{cao2023efficient}), multi-stage quantization (RQ-VAE \cite{yu2022vectorquantized}, MoVQ \cite{zheng2022movq}), and lookup-free schemes (MAGVIT-v2 \cite{yu2024language}, FSQ \cite{mentzer2024finite}) for improved efficiency and expressiveness.
%Subsequent improvements included adversarial training \cite{yu2022vectorquantized}, transformer architectures \citep{yu2022vectorquantized}, better efficiency \cite{cao2023efficient}, multi-stage quantization \cite{yu2022vectorquantized, zheng2022movq}, and lookup-free schemes %for improved efficiency and expressiveness 
%\cite{yu2024language, mentzer2024finite}.
later improved and extended by several works \citep{yu2022vectorquantized, yu2024language, mentzer2024finite}.
While early methods encode images into 2D grids of latents, recent works have developed 1D tokenizers that return a sequence of tokens without spatial structure \cite{yu2024titok, miwa2025one, bachmann2025flextok, ma2025unitok}.
\titok \cite{yu2024titok} demonstrates that as few as 32 tokens can produce high-quality reconstructions.
% \unitok \cite{ma2025unitok} focuses on building representations effective for both generation and understanding by integrating reconstruction and CLIP supervision during training. \unitok introduces multi-codebook quantization and attention-based projection, enhancing latent space expressivity and enabling tokenizers to produce representations that generalize well across both generative and discriminative tasks.
\unitok \cite{ma2025unitok} improves generalization in generative and discriminative tasks by introducing multi-codebook quantization while integrating reconstruction and CLIP supervision. %and attention-based projection to enhance latent space expressivity
% While TiTok and UniTok tokenizers have a fixed number of tokens, other approaches such as One-D-Piece \cite{miwa2025one} and FlexTok \cite{bachmann2025flextok} allow varying the number of tokens to trade-off compression rate and reconstruction quality at inference time.
Other approaches, such as One-D-Piece \cite{miwa2025one} and FlexTok \cite{bachmann2025flextok}, allow varying the number of tokens to balance compression and reconstruction quality at inference.

\begin{figure*}[t]
    \centering
    \includegraphics[width=.9\linewidth]{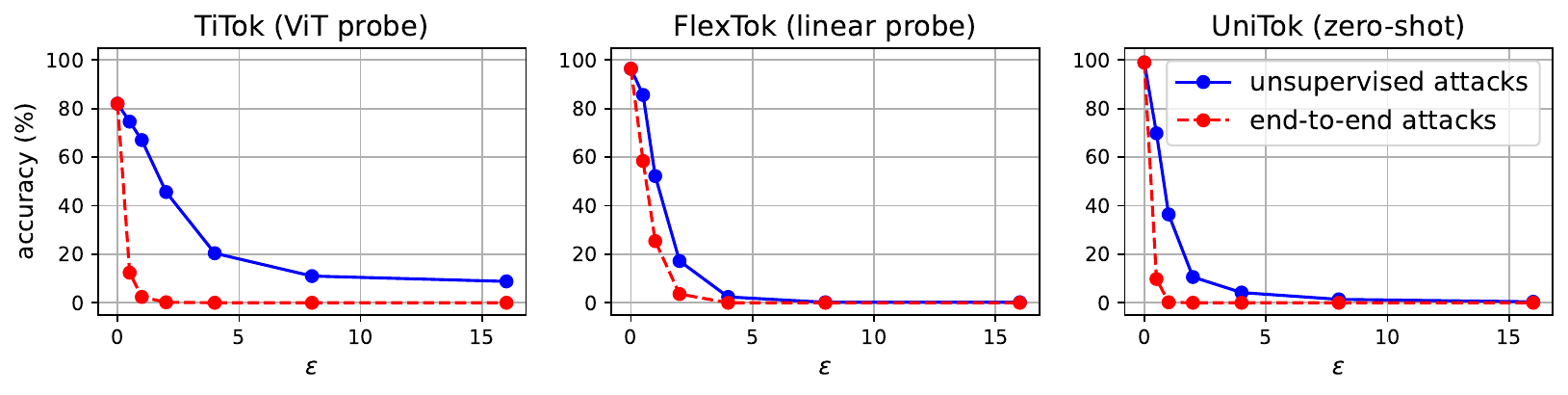}
    % \vspace{-3mm} 
    \caption{\textbf{Unsupervised vs supervised adversarial attacks for classification.}
    We report the robust accuracy when varying the perturbation radius $\epsilon$ (scaled to [0, 255]) for three classifiers on Imagenette: \titok with ViT probing (left), \flextok with linear probing (middle), and zero-shot \unitok (right).
    In most cases, our unsupervised attacks (blue curves), which target only the image tokenizer and do not need label information, perform close to end-to-end supervised attacks (red), which target the entire classifier and use labels. For small $\epsilon$, our unsupervised attacks are slightly worse than the supervised ones.
    Both attacks are optimized with 100 iterations of APGD on 500 images.\looseness-1
    }\label{fig:sup-vs-unsup}
%\vspace{-1em}
\end{figure*}

%\fra{can be skipped} 
% \myparagraph{Discrete image tokenizers in downstream tasks.}
% While their corresponding de-tokenizers are commonly used for image generation in frameworks such as MaskGIT \cite{chang2022maskgit}, discrete image tokenizers have been recently used as components of models for downstream tasks. First, the features extracted by the tokenizers can be used for transfer learning in classification tasks, either zero-shot (\unitok \cite{ma2025unitok}) or with linear probing (\flextok \cite{bachmann2025flextok}). Second, encoder-only (\fuselip \cite{schlarmann2025fuselip}) and encoder-decoder models \cite{mizrahi20234m, bachmann20244m21, lu2022unified} leverage discrete image tokenizers to extend masked modeling losses to visual \rev{inputs}, beyond language data. Finally, multiple early-fusion autoregressive models, such as Chameleon \cite{team2024chameleon}, Show-o \cite{xie2025showo}, UniTok-MLLM \cite{ma2025unitok}, SelfTok \cite{wang2025selftok} use discrete tokenizers to convert input images into a sequence of tokens from a fixed codebook, providing a setup similar to natural language.\looseness-1

\myparagraph{Adversarial robustness of image encoders.}
% \fra{maybe add some citations e.g. visual jailbreak}
Adversarial robustness to $\ell_p$-bounded attacks has been extensively studied for a variety of vision tasks, such as image classification \cite{croce2020RobustBench}, semantic segmentation \cite{croce2024towards}, object detection \cite{li2025importance}, with the development of many algorithms to both generate adversarial perturbations \cite{carlini2017towards, Croce2020Autoattack} and improve the robustness to deep learning models \cite{Madry2018AT, zhang2019theoretically}.
With the development of foundation multimodal models such as CLIP \cite{radford2021clip} and SAM \cite{kirillov2023segment}, those approaches have been adapted to modern scenarios.
Moreover, new techniques have been designed to attack multimodal autoregressive %large language models 
LLMs like LLaVA \cite{li2024llava} by perturbing the input images, which allows an attacker to control the model output %generated by the model 
\cite{schlarmann2023adversarial, qi2024visual}.
These works reveal the persisting vulnerability of image encoders to adversarial perturbations, both in isolation and when used as part of more complex systems  \cite{bhagwatkar-etal-2024-improving}.
Therefore, a few works have proposed methods to improve their robustness, %typically leveraging different forms of adversarial training \cite{Madry2018AT}.
%for example, 
such as \citet{Mao2022UnderstandingZAtecoa, schlarmann2024robust} who fine-tuned the vision encoder of CLIP with variants of adversarial training \cite{Madry2018AT}.
% \fra{there are more} 
However, to our knowledge, the robustness of discrete image tokenizers has not been studied so far.
% \vspace{-1em}

%\section{Background and Related Work}

\section{Testing and Improving the Adversarial Robustness of Image Tokenizers via Unsupervised Attacks
%Unsupervised Adversarial Attacks against Image Tokenizers
}\label{sec:method}

%In the following,
We first provide a background on discrete image tokenizers, then introduce our unsupervised attacks as well as experiments supporting their effectiveness,
%Finally, we present how these can be leveraged to improve the robustness of tokenizers.
and finally discuss how to leverage them to improve the robustness of tokenizers.

\noindent \textbf{Background.}
%Continuous-space image encoders transform an image to a sequence of embeddings or tokens $\phi: I \rightarrow \mathbb{R}^{T\times d}$, where $T$ is the number of tokens and $d$ is the latent dimension.
Discrete image tokenizers extract a sequence of $T$ $d$-dimensional latent vectors from an input image.
Concretely, they first leverage an encoder $\phi: I \rightarrow \mathbb{R}^{T\times d}$, typically a CNN or vision transformer, that maps an image $x$ to $T$ embeddings, $\,\phi(x) = \{h_i\}_{i=1}^T \in\mathbb{R}^{T \times d}$.
Then, a vector quantizer with a learned codebook $\mathcal{C}=\{e_k\in\mathbb{R}^d\}_{k=1}^K$ replaces each pre-quantization embedding $h_i\in\mathbb{R}^d$ (at location $i$) with its nearest code,
%\begin{align}
%q_i=\underset{k\in[K]}{\argmin}\|h_i-e_k\|_2.
%\end{align}
i.e., $q_i=\argmin_{k\in[K]}\norm{h_i-e_k}_2$.
This yields a discrete index map $\{q_i\}_{i=1}^{T}$.
Unlike continuous encoders that output continuous space embeddings and remain differentiable everywhere, nearest-neighbor quantization is piecewise constant since all embeddings with a particular nearest code are mapped to it, creating discretized cells in the continuous latent space. 
Hence, even small shifts at the boundaries of these cells can %lead to a new
alter nearest code and index.
%In practice, the 
Encoder and codebook are learned jointly using straight-through updates. 
%Unlike continuous tokenizers, where only $d$ influences the representational capacity, here, even the codebook size $K$ is crucial.
% and both the latent dimension $d$ and the codebook size $K$ influence the representational capacity.

\subsection{Unsupervised attacks via embedding distortion}
%\myparagraph{Unsupervised attacks via embedding distortion.}

Pre-trained image tokenizers are used as plug-ins in complex systems for multiple downstream tasks. %(e.g., reconstruction, classification)
Therefore, we aim to develop attacks that can be effective regardless of the downstream application.
Since the representation of an image $x$ used by any downstream model is determined by the codebook tokens output by the tokenizer, we expect that changing the tokens will lead to an uninformative or distorted encoding of $x$.
While we could aim to directly change the index sequence returned by the tokenizer, the indices themselves carry no (or minimal) semantic or perceptual information, and the resulting problem is not differentiable.
Hence, we propose an attack in the pre-quantization embedding space of discrete image tokenizers.
Specifically, we maximize the $\ell_2$-distance between the embeddings of the clean and perturbed images, similar to existing attacks on continuous encoders \cite{croce2024segment, schlarmann2024robust}, i.e., we aim to solve
\begin{align}
%\underset{\|\delta\|_\infty \leq \epsilon}{\max}\left\| \phi(x + \delta) - \phi(x) \right\|_2,
\max_{\norm{\delta}_p \leq \epsilon} \; \sum_{i=1}^T \norm{h_i(x + \delta) - h_i(x)}_2^2,
\label{eq:unsupervised-objective}
\end{align}
%where $\phi(x) = \{h_i\}_{i=1}^T \in \mathbb{R}^{T \times d}$
where $h_i(\cdot)\in \mathbb{R}^{d}$ is the $i$-th pre-quantization embedding produced by the image encoder, and $\delta$ is the adversarial perturbation constrained by an $\ell_p$-norm bound $\epsilon$.
% \fra{check notation, check loss formulation}
Intuitively, our approach aims to shift the encoder output sufficiently far in latent space to alter the resulting quantized codes, thereby corrupting any downstream application of the tokens.
We remark that our objective directly manipulates the encoder output (prior to vector quantization), without requiring access to the codebook or any information (e.g., class labels) of the downstream applications.

%Unlike standard classification attacks that rely on downstream task losses, our objective directly manipulates the encoder output $\phi(x)$ prior to vector quantization.

\begin{figure*}
\centering \small
\tabcolsep=2pt
    \begin{tabular}{ *{3}{C{15mm}} | *{3}{C{15mm}} | *{3}{C{15mm}} }
    \multicolumn{3}{c|}
    {\textbf{\titok}} & \multicolumn{3}{c|}
    {\textbf{\flextok}} & \multicolumn{3}{c}
    {\textbf{\unitok}} \\[1.1mm]
    clean & $\nicefrac{4}{255}$ & $\nicefrac{8}{255}$ & clean & $\nicefrac{4}{255}$ & $\nicefrac{8}{255}$ & clean & $\nicefrac{4}{255}$ & $\nicefrac{8}{255}$ \\

    \multicolumn{3}{c|}{\includegraphics[width=.6\columnwidth]{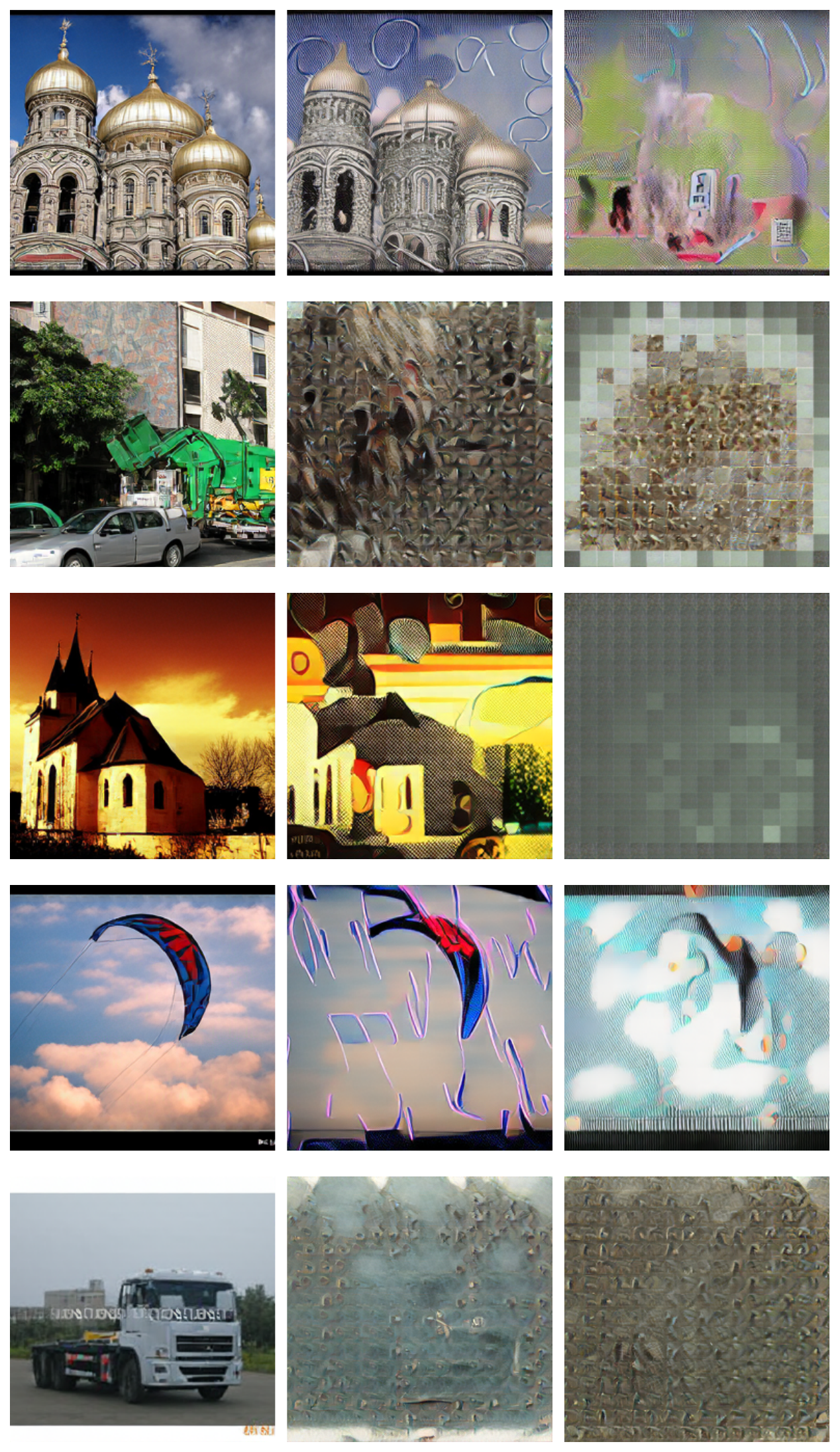}} 
    
    &
    \multicolumn{3}{c|}{\includegraphics[width=.6\columnwidth]{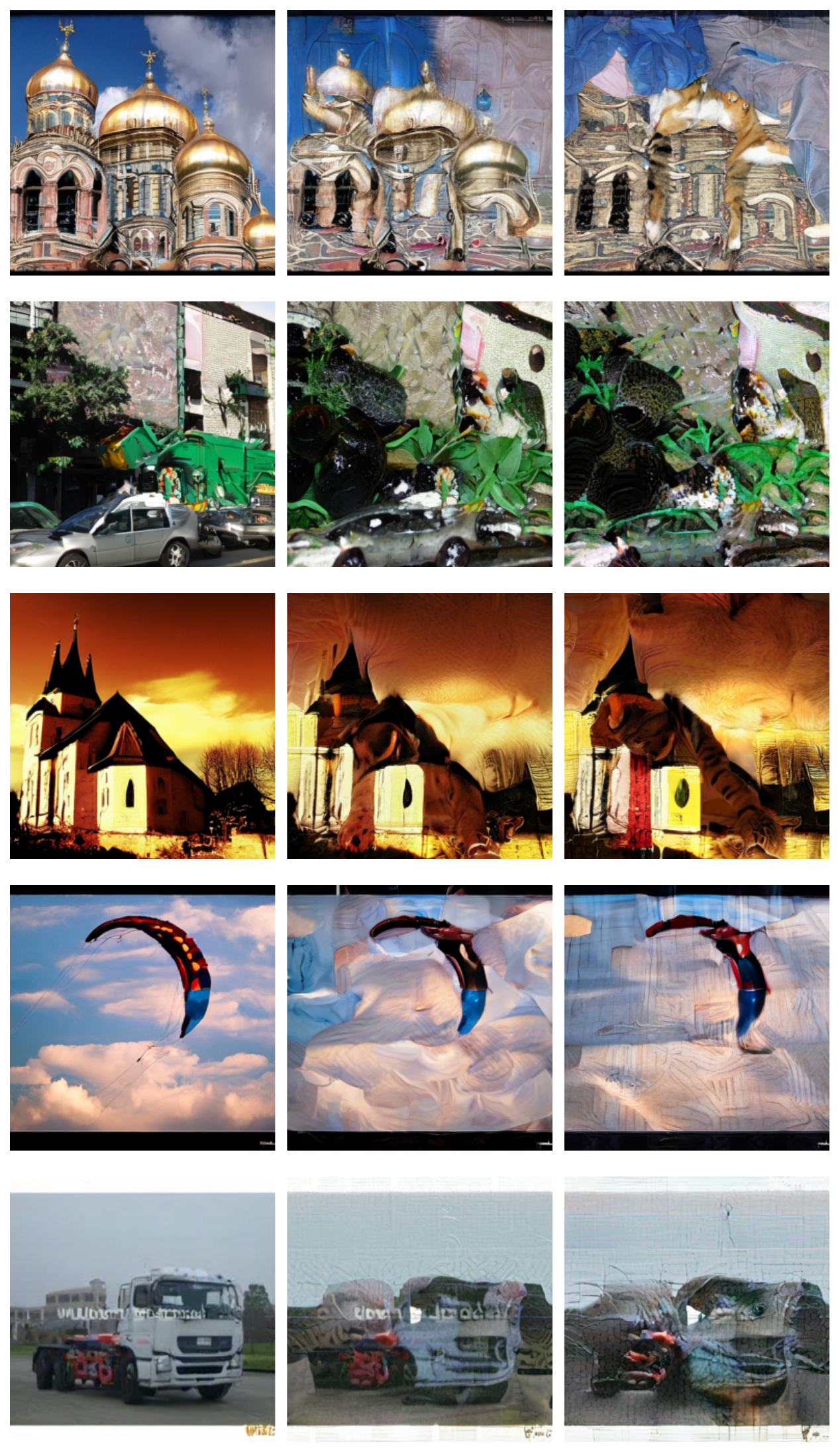}}

    &
    \multicolumn{3}{c}{\includegraphics[width=.6\columnwidth]{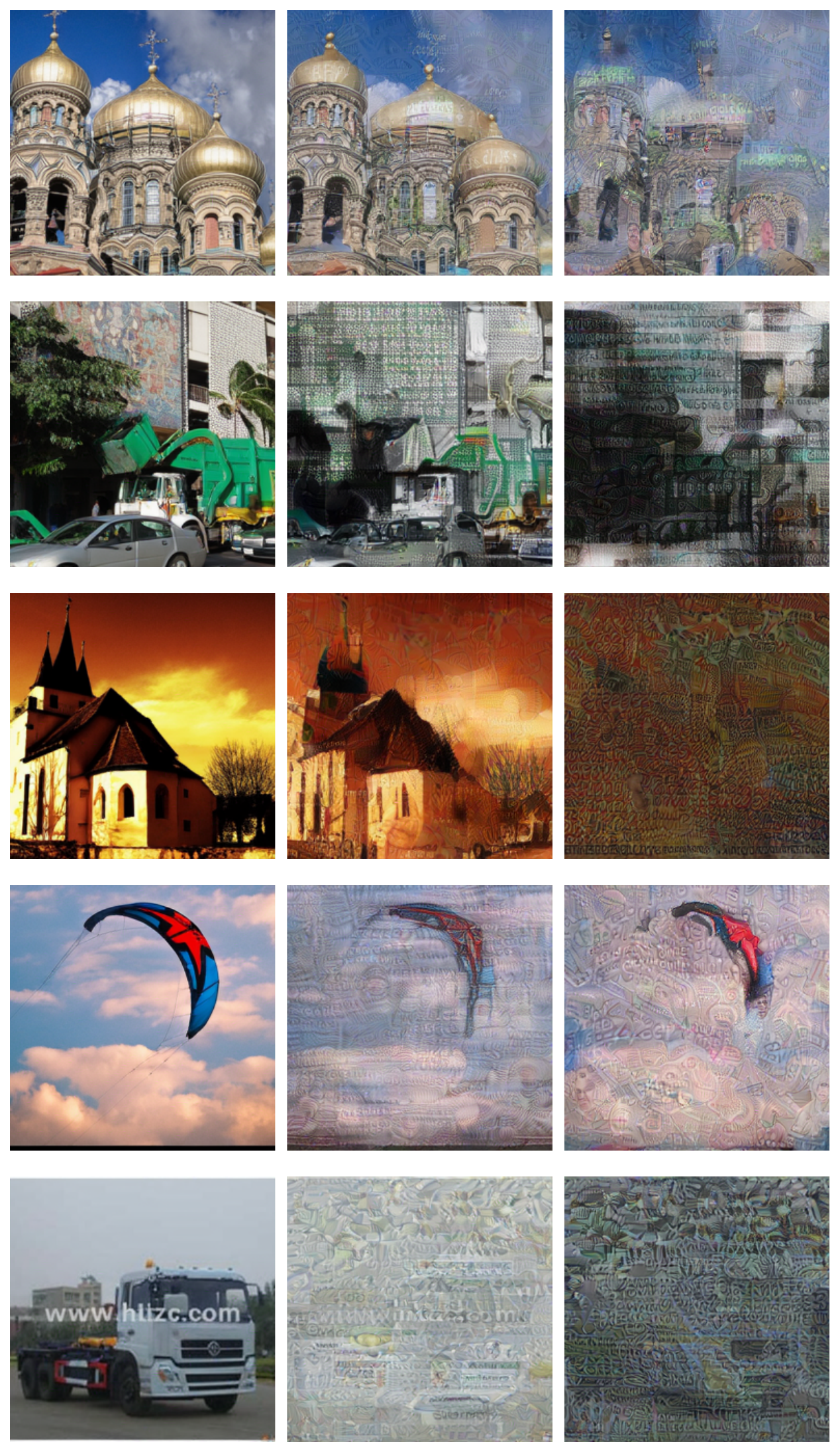}}
         
    \end{tabular}
    % \vspace{-2mm}
    \caption{%Reconstruction examples, 2.5k iterations
    \textbf{Reconstruction of unsupervised attacks.}
    For each tokenizer, we show the reconstruction (given by the corresponding de-tokization models) of the clean images and adversarial images computed by unsupervised attacks at $\epsilon=\nicefrac{4}{255}, \nicefrac{8}{255}$ with 2500 \rev{steps of APGD}.
    The perturbed inputs affect the reconstruction differently depending on the tokenizer, with \titok yielding the most distorted decoded images while \flextok being most robust, with still clearly recognizable subjects.
    }\label{fig:reconstruction}
    \vspace{-1em}
\end{figure*}

\myparagraph{Effectiveness of unsupervised attacks.} %\fra{write}
To assess the effectiveness of our unsupervised attack, we first compare it to end-to-end supervised attacks on image classification.
We obtain three diverse classifiers based on discrete tokenizers on the Imagenette dataset \cite{imagenette}: we train a ViT probe on the quantized features of \titok-BL128 \cite{yu2024titok}, a linear probe on FlexTok \cite{bachmann2025flextok}, and zero-shot \unitok \cite{ma2025unitok}.
We compare unsupervised attacks, where APGD \cite{Croce2020Autoattack} is used to optimize Eq.~\eqref{eq:unsupervised-objective}, to a standard end-to-end attack (APGD on the cross entropy loss) which targets the entire classifier and requires label information, both with 100 iterations.
Fig.~\ref{fig:sup-vs-unsup} shows the robust accuracy of the classifiers when varying the radius $\epsilon$ of the $\ell_\infty$-bounded attacks. 
Notably, our unsupervised attacks (blue curves) often perform close to the end-to-end supervised attacks (red), lagging behind particularly for small $\epsilon$ values.
This also shows that even at limited computational cost (only 100 iterations), unsupervised attacks are effective in fooling the classifiers, and hence are useful for adversarial training (see Sec.~\ref{sec:method}).
Finally, we see that the classifiers built on discrete tokenizers are highly vulnerable to adversarial perturbations.

\myparagraph{Reconstruction of adversarial images.}
A unique aspect of our setup is that we can use the image de-tokenizers to reconstruct the adversarial images derived by unsupervised attacks.
We qualitatively estimate how distorting the features extracted by the tokenizer's encoder impacts the reconstruction.
In Fig.~\ref{fig:reconstruction}, we provide several examples of reconstruction of unsupervised attacks (2500 iterations of APGD) at different $\epsilon$.
Despite these attacks being similarly effective on classification (as mentioned above), they affect reconstruction differently depending on the tokenizer, with \titok yielding the most distorted reconstructed images while \flextok being the most robust with still clearly recognizable subjects.
This hints to some structural difference among image tokenization approaches which might impact robustness, presenting an interesting direction for future work.\looseness-1

\subsection{Robust tokenizers via unsupervised adversarial fine-tuning}

A common approach to improve the robustness of neural networks is adversarial training \cite{Madry2018AT}, where adversarial perturbations are generated on the fly during training and used to augment the original training set.
In our case, we want to make the image tokenization invariant to adversarial perturbations while preserving its effectiveness for downstream applications.
Inspired by the approach of \citet{schlarmann2024robust} for improving the robustness of CLIP's image encoder, we propose adversarially fine-tuning the image tokenizer on our unsupervised attack.
This yields 
\begin{align}
%\underset{\|\delta\|_\infty \leq \epsilon}{\max}\left\| \phi(x + \delta) - \phi(x) \right\|_2,
\min_\theta \; \frac{1}{|\mathcal{D}|} \sum_{x \in \mathcal{D}} \,\max_{\norm{\delta}_p \leq \epsilon} \; \sum_{i=1}^T \, \norm{h_i^\theta(x + \delta) - h_i^{\theta_\textrm{orig}}(x)}_2^2,
\label{eq:unsupervised-at}
\end{align}
as the training objective, where $\theta$ are the parameters of the fine-tuned tokenizer, $\theta_\textrm{orig}$ the parameters of the original tokenizer (before fine-tuning), and $\mathcal{D}$ is the training dataset.
This loss pushes the fine-tuned tokenizer to yield consistent embeddings in an $\ell_p$-ball of radius $\epsilon$ around the embeddings given by the original tokenizer.
Therefore, the fine-tuned tokenizers could readily replace the original tokenizers in downstream tasks without hurting clean performance while improving robustness.
In principle, the quantization steps of discrete tokenizers make it possible to preserve the downstream performance even without exactly solving Eq.~\eqref{eq:unsupervised-at}, since sufficiently close embeddings lead to the same codebook indices.
% \fra{maybe this can be written somewhere else} 
Compared to task-specific end-to-end supervised adversarial training, our task-agnostic approach has several advantages: \textit{(i)} it yields tokenizers which can be deployed across multiple downstream applications, \textit{(ii)} can leverage virtually any image dataset, even beyond that used for pre-training (as we show in Sec.~\ref{sec:analyses}) since it does not need label information, and \textit{(iii)} fine-tunes only a subset of the parameters, i.e., those of the tokenizer's encoder, substantially lowering the computational cost (as we show in Sec.~\ref{sec:analyses}).\looseness-1
%\end{itemize}

% We implement this attack using the APGD optimization method, adapted to our custom objective. The attack is run for 50 iterative steps, and gradients are computed with respect to the distance-based loss rather than any classification or reconstruction loss. 

%Since the vector quantization step is sensitive to small shifts near code boundaries, even modest changes in the pre-quantized embeddings can induce significant changes in the discrete token indices. Therefore, this attack is well-suited to probing the robustness of discrete tokenizers, regardless of the downstream task.

%Compared to standard end-to-end attacks that maximize a supervised loss (e.g., cross-entropy), our pre-quantization objective offers \colorbox{yellow}{X} practical advantages. \emph{(i) Task/label agnosticism:} it requires no labels or task-specific heads, so the same attack applies uniformly to reconstruction, generative, and discriminative pipelines that consume discrete tokens.

\section{Experiments}

%\subsection{Setup}

%\textbf{Tokenizers.}
In the following, we present results for the original and our adversarially fine-tuned tokenizers across tasks and datasets.
In detail, we experiment with two vector-quantized image tokenizers,  TiTok-BL128 and UniTok, since both, unlike \flextok, are part of larger models and can be tested on multiple downstream tasks.
In addition, we report the results of the smaller TiTok-S128 in App.~\ref{sec:additional_experiments_results}. 
%
%\textbf{Adversarial training details.}
To obtain the robust versions, we adversarially fine-tune only the encoder of each tokenizer, while keeping the codebook, downstream decoders, LLMs or any other components frozen. 
We fine-tune for one epoch under $\ell_\infty$-bounded perturbations with our unsupervised embedding-space attack (10 steps of APGD), see Sec.~\ref{sec:method}, with perturbation radii $\epsilon \in \{\nicefrac{4}{255}, \nicefrac{8}{255}, \nicefrac{12}{255}, \nicefrac{16}{255}\}$, on \imnet-1k.
More details on the setup are in App.~\ref{sec:exp_details}.
Remarkably, despite adversarially fine-tuning only on ImageNet-1k, the resulting robust tokenizers exhibit strong robustness across a wide range of downstream datasets and tasks, highlighting their ability to transfer robustness beyond the training domain.

\begin{table*}[t]
\centering
\small
\setlength{\tabcolsep}{4.5pt}
\caption{\textbf{Evaluation of \fuselip on image classification and multimodal retrieval.}  We report the clean and robust accuracy (\%) %on Imagenette, Caltech101, OI-CROP and OI-POS 
under %no attack (clean) and 
\(\ell_\infty\)-bounded perturbations with \(\epsilon\in\{2/255,4/255\}\) for original and robust tokenizers trained on different radii. %The rightmost block reports averages across datasets.
}
\label{tab:titok_b_supervised}
%\vspace{-2mm}
\begin{tabular}{@{}l|ccc|ccc|ccc|ccc|ccc@{}}
\toprule
\multirow{2}{*}{Tokenizer} 
& \multicolumn{3}{c|}{\textbf{Imagenette}}
& \multicolumn{3}{c|}{\textbf{Caltech101}}
& \multicolumn{3}{c|}{\textbf{OI-Crop}}
& \multicolumn{3}{c|}{\textbf{OI-Pos}}
& \multicolumn{3}{c}{\textbf{Average}} \\
% \cmidrule(lr){1-1}
% \cmidrule(lr){2-4}\cmidrule(lr){5-7}\cmidrule(lr){8-10}\cmidrule(l){11-13}
\cmidrule{2-16}
% &   clean & \multicolumn{2}{c|}{$\ell_\infty$}
% & clean & \multicolumn{2}{c|}{$\ell_\infty$}
% & clean & \multicolumn{2}{c|}{$\ell_\infty$}
% & clean & \multicolumn{2}{c|}{$\ell_\infty$}
% & clean & \multicolumn{2}{c}{$\ell_\infty$} \\
% \cmidrule{3-4} \cmidrule{6-7} \cmidrule{9-10} \cmidrule{12-13}\cmidrule{15-16}
% &    &  $\nicefrac{2}{255}$ &  $\nicefrac{4}{255}$
% &  & $\nicefrac{2}{255}$ &  $\nicefrac{4}{255}$
% &  & $\nicefrac{2}{255}$ &  $\nicefrac{4}{255}$
% &  & $\nicefrac{2}{255}$ &  $\nicefrac{4}{255}$
% &  & $\nicefrac{2}{255}$ &  $\nicefrac{4}{255}$ \\
& clean &  $\nicefrac{2}{255}$ &  $\nicefrac{4}{255}$ & clean &  $\nicefrac{2}{255}$ &  $\nicefrac{4}{255}$& clean &  $\nicefrac{2}{255}$ &  $\nicefrac{4}{255}$ & clean &  $\nicefrac{2}{255}$ &  $\nicefrac{4}{255}$ & clean &  $\nicefrac{2}{255}$ &  $\nicefrac{4}{255}$ \\
\midrule
 % clean              & 93.6 &  2.6 &  0.0 & 74.4 &  0.6 &  0.0 & 71.8 &  7.4 &  0.8 & 69.2 &  5.4 &  1.4 & 77.3 &  4.0 &  0.6 \\
 % $\text{AT}^{\nicefrac{4}{255}}$    & 92.2	& 63.2	& 36.0	& 73.0 & 48.2 & 19.8 & 65.4	&49.4	&26.6 &	67.2&	47.6	&23.8&	74.5	&52.1	&26.6\\
 % $\text{AT}^{\nicefrac{8}{255}}$     & 89.0 &	67.8 &	46.4 &	73.2 &	48.6 &	32.4 &	59.6 &	47.8 &	35.0 &	65.2 &	53.6 &	34.4 &	71.8 &	54.5 &	37.1 \\
% $\text{AT}^{\nicefrac{12}{255}}$   & 86.4 & 	68.4	&	50.4 &67.6	& 47.6 &	36.0 &	54.4 &	49.0 &	37.2 &	62.6 &	49.4 &	34.8 &	67.8 &	53.6 &	39.6 \\
% $\text{AT}^{\nicefrac{16}{255}}$   & 81.6 &	62.2 &	47.8 &	59.6 &	46.2 &	35.4 &	50.0 &	46.2 &	33.4 &	56.8 &	47.6 &	37.4 &	62.0 &	50.6 &	38.5\\

% 10 steps results below
% clean & 93.6 & 2.6 & 0.00 & 74.40 & 0.6 & 0.00 & 71.80 & 7.40 & 0.80 & 69.20 & 5.40 & 1.40 & 77.25 & 4.00 & 0.55 \\
% $\text{AT}^{\nicefrac{4}{255}}$ & 91.80 & 63.6 & 36.6 & 73.00 & 48.20 & 20.80 & 66.20 & 50.6 & 26.00 & 67.20 & 46.00 & 24.6 & 74.55 & 52.10 & 27.00 \\
% $\text{AT}^{\nicefrac{8}{255}}$ & 89.6 & 69.00 & 48.80 & 72.40 & 51.6 & 32.80 & 62.00 & 48.80 & 35.80 & 64.80 & 51.20 & 35.6 & 72.20 & 55.15 & 38.25 \\
% $\text{AT}^{\nicefrac{12}{255}}$ & 87.00 & 71.40 & 51.00 & 67.6 & 51.20 & 36.80 & 56.20 & 49.00 & 36.80 & 61.6 & 49.6 & 35.20 & 68.10 & 55.30 & 39.95 \\
% $\text{AT}^{\nicefrac{16}{255}}$ & 83.40 & 66.6 & 50.00 & 61.20 & 47.6 & 37.40 & 50.00 & 47.20 & 35.80 & 59.40 & 48.80 & 39.20 & 63.50 & 52.55 & 40.6\\

%
clean & 93.6 & 2.6 & 0.0 & 74.4 & 0.6 & 0.0 & 71.8 & 7.4 & 0.8 & 69.2 & 5.4 & 1.4 & 77.3 & 4.0 & 0.6 \\
$\text{AT}^{\nicefrac{4}{255}}$ & 91.8 & 63.6 & 36.6 & 73.0 & 48.2 & 20.8 & 66.2 & 50.6 & 26.0 & 67.2 & 46.0 & 24.6 & 74.6 & 52.1 & 27.0 \\
$\text{AT}^{\nicefrac{8}{255}}$ & 89.6 & 69.0 & 48.8 & 72.4 & 51.6 & 32.8 & 62.0 & 48.8 & 35.8 & 64.8 & 51.2 & 35.6 & 72.2 & 55.2 & 38.3 \\
$\text{AT}^{\nicefrac{12}{255}}$ & 87.0 & 71.4 & 51.0 & 67.6 & 51.2 & 36.8 & 56.2 & 49.0 & 36.8 & 61.6 & 49.6 & 35.2 & 68.1 & 55.3 & 40.0 \\
$\text{AT}^{\nicefrac{16}{255}}$ & 83.4 & 66.6 & 50.0 & 61.2 & 47.6 & 37.4 & 50.0 & 47.2 & 35.8 & 59.4 & 48.8 & 39.2 & 63.5 & 52.6 & 40.6 \\
\bottomrule
\end{tabular}
\end{table*}

\begin{table*}[t]
\centering
\small
\caption{\textbf{Evaluation of UniTok on image classification.}  For each dataset, we report the accuracy (\%) under no attack (clean) and \(\ell_\infty\)-bounded perturbations with \(\epsilon\in\{2/255,4/255\}\) for original and robust tokenizers trained on different radii. %The rightmost block reports averages across datasets.
}
%\vspace{-2mm}
\setlength{\tabcolsep}{4.5pt}
\begin{tabular}{@{}L{15mm} %|ccc|ccc|ccc|ccc
*{4}{|*{3}{C{8mm}}} @{}}
\toprule
\multirow{2}{*}{Tokenizer} 
& \multicolumn{3}{c|}{\textbf{Imagenette}}
& \multicolumn{3}{c|}{\textbf{Caltech101}}
& \multicolumn{3}{c|}{\textbf{ImageNet}}
& \multicolumn{3}{c}{\textbf{Average}} \\
% \cmidrule(lr){1-1}
% \cmidrule(lr){2-4}\cmidrule(lr){5-7}\cmidrule(lr){8-10}\cmidrule(l){11-13}
\cmidrule{2-13}
% &   clean & \multicolumn{2}{c|}{$\ell_\infty$}
% & clean & \multicolumn{2}{c|}{$\ell_\infty$}
% & clean & \multicolumn{2}{c|}{$\ell_\infty$}
% % & clean & \multicolumn{2}{c|}{$\ell_\infty$}
% & clean & \multicolumn{2}{c}{$\ell_\infty$} \\
% \cmidrule{3-4} \cmidrule{6-7} \cmidrule{9-10} \cmidrule{12-13}
% &    &  $\nicefrac{2}{255}$ &  $\nicefrac{4}{255}$
% &  & $\nicefrac{2}{255}$ &  $\nicefrac{4}{255}$
% &  & $\nicefrac{2}{255}$ &  $\nicefrac{4}{255}$
% % &  & $\nicefrac{2}{255}$ &  $\nicefrac{4}{255}$
% &  & $\nicefrac{2}{255}$ &  $\nicefrac{4}{255}$ \\
%
& clean &  $\nicefrac{2}{255}$ &  $\nicefrac{4}{255}$ & clean &  $\nicefrac{2}{255}$ &  $\nicefrac{4}{255}$& clean &  $\nicefrac{2}{255}$ &  $\nicefrac{4}{255}$ & clean &  $\nicefrac{2}{255}$ &  $\nicefrac{4}{255}$\\
\midrule
 %clean              & 99.21 & 0.00 & 0.00 & 85.71 & 0.00 & 0.00 & 67.26 & 0.00 & 0.00 & 84.06 & 0.00 & 0.00 \\
%  $\text{AT}^{\nicefrac{4}{255}}$    & 99.0 &	91.3 &	75.2 &	81.3 &	51.8 &	23.2  &	66.5 &	33.5 &	14.1 &	82.3 &	58.9 &	37.5\\
%  $\text{AT}^{\nicefrac{8}{255}}$     & 97.4 &	92.1 &	81.9 &	77.8 &	58.5 &	42.9 &	58.5 &	37.5 &	24.6 &	77.9 &	62.7 &	49.8 \\
% $\text{AT}^{\nicefrac{12}{255}}$   & 95.6 &	91.1 &	82.3 &	72.2 &	56.7 &	46.6 &	50.2 &	35.9 &	25.0 &	72.7 &	61.2 &	51.3 \\
% $\text{AT}^{\nicefrac{16}{255}}$   & 91.5 &	87.5 &	79.2 &	65.1 &	 54.2 &	43.8 &	41.7 &	30.6 &	23.0 &	66.1 &	57.4 &	48.7\\

%10 steps results below
% $\text{AT}^{\nicefrac{4}{255}}$ & 99.21 & 92.06 & 75.00 & 81.15 & 56.94 & 22.42 & 66.87 & 31.94 & 10.52 & 82.41 & 60.32 & 35.98 \\
% $\text{AT}^{\nicefrac{8}{255}}$ & 97.82 & 91.47 & 82.74 & 77.38 & 63.49 & 43.85 & 58.33 & 40.28 & 23.61 & 77.84 & 65.08 & 50.07 \\
% $\text{AT}^{\nicefrac{12}{255}}$ & 95.63 & 88.69 & 81.35 & 72.42 & 60.12 & 47.62 & 50.40 & 36.51 & 25.6 & 72.82 & 61.77 & 51.52 \\
% $\text{AT}^{\nicefrac{16}{255}}$ & 92.66 & 86.31 & 79.56 & 65.28 & 57.54 & 44.64 & 42.26 & 32.14 & 23.61 & 66.73 & 58.66 & 49.27\\
%\midrule
%
clean & 99.2 & 0.0 & 0.0 & 85.7 & 0.0 & 0.0 & 67.3 & 0.0 & 0.0 & 84.1 & 0.0 & 0.0 \\
$\text{AT}^{\nicefrac{4}{255}}$ & 99.2 & 92.1 & 75.0 & 81.2 & 56.9 & 22.4 & 66.9 & 31.9 & 10.5 & 82.4 & 60.3 & 36.0 \\
$\text{AT}^{\nicefrac{8}{255}}$ & 97.8 & 91.5 & 82.7 & 77.4 & 63.5 & 43.9 & 58.3 & 40.3 & 23.6 & 77.8 & 65.1 & 50.1 \\
$\text{AT}^{\nicefrac{12}{255}}$ & 95.6 & 88.7 & 81.4 & 72.4 & 60.1 & 47.6 & 50.4 & 36.5 & 25.6 & 72.8 & 61.8 & 51.5 \\
$\text{AT}^{\nicefrac{16}{255}}$ & 92.7 & 86.3 & 79.6 & 65.3 & 57.5 & 44.6 & 42.3 & 32.1 & 23.6 & 66.7 & 58.7 & 49.3 \\
\bottomrule
\end{tabular}

\label{tab:unitok_classification_supervised}
% \vspace{-3mm}
\end{table*}

\begin{table*}[t]
\centering
\small
\caption{\textbf{UniTok-MLLM on VQA.}  We report the accuracy (\%) on VQAv2, OK-VQA, GQA and their average under no attack (clean) and \(\ell_\infty\)-bounded perturbations with \(\epsilon\in\{2/255,4/255\}\) for original and robust tokenizers trained on different radii. %The rightmost block reports averages across datasets.
}
%\vspace{-2mm}
\setlength{\tabcolsep}{6.5pt}
\begin{tabular}{@{}l|ccc|ccc|ccc|ccc@{}}
\toprule
\multirow{2}{*}{Tokenizer} 
& \multicolumn{3}{c|}{\textbf{VQAv2}}
& \multicolumn{3}{c|}{\textbf{OK-VQA}}
& \multicolumn{3}{c|}{\textbf{GQA}}
& \multicolumn{3}{c}{\textbf{Average}} \\
% \cmidrule(lr){1-1}
% \cmidrule(lr){2-4}\cmidrule(lr){5-7}\cmidrule(lr){8-10}\cmidrule(l){11-13}
\cmidrule{2-13}
% &   clean & \multicolumn{2}{c|}{$\ell_\infty$}
% & clean & \multicolumn{2}{c|}{$\ell_\infty$}
% & clean & \multicolumn{2}{c|}{$\ell_\infty$}
% % & clean & \multicolumn{2}{c|}{$\ell_\infty$}
% & clean & \multicolumn{2}{c}{$\ell_\infty$} \\
% \cmidrule{3-4} \cmidrule{6-7} \cmidrule{9-10} \cmidrule{12-13}
% &    &  $\nicefrac{2}{255}$ &  $\nicefrac{4}{255}$
% &  & $\nicefrac{2}{255}$ &  $\nicefrac{4}{255}$
% &  & $\nicefrac{2}{255}$ &  $\nicefrac{4}{255}$
% % &  & $\nicefrac{2}{255}$ &  $\nicefrac{4}{255}$
% &  & $\nicefrac{2}{255}$ &  $\nicefrac{4}{255}$ \\
%
& clean &  $\nicefrac{2}{255}$ &  $\nicefrac{4}{255}$ & clean &  $\nicefrac{2}{255}$ &  $\nicefrac{4}{255}$& clean &  $\nicefrac{2}{255}$ &  $\nicefrac{4}{255}$ & clean &  $\nicefrac{2}{255}$ &  $\nicefrac{4}{255}$\\
\midrule
% clean        &73.20 & 14.42 & 8.46 & 59.56 & 2.12 & 0.84 & 68.00 & 13.40 & 10.00 & 66.92 & 9.98 & 6.43    \\
%  $\text{AT}^{\nicefrac{4}{255}}$ &     67.1 &	45.6 &	32.8 &	52.9 &	36.2 &	23.6 &	66.6 &	46.0 &	33.0 &	62.2 &	42.6 &	29.8\\
%  $\text{AT}^{\nicefrac{8}{255}}$  &    64.0 &	49.3 &	40.0 &	48.6 &	38.2 &	30.8 &	66.2 &	48.2 &	39.2 &	59.6 &	45.2 &	36.7 \\
% $\text{AT}^{\nicefrac{12}{255}}$  &  60.7	& 50.1	& 41.5	& 47.6	& 37.2	&31.0 &	64.0 &	49.6&	39.6&	57.4 &	45.6 &	37.4 \\
% $\text{AT}^{\nicefrac{16}{255}}$   &  58.9 &	47.9 &	40.2 &	45.1 &	37.4 &	30.4 &	61.2 &	50.2 &	40.6 &	55.1 &	45.1 &	37.0\\

%10 steps results below
% $\text{AT}^{\nicefrac{4}{255}}$ &     67.36 & 50.52 & 44.36 & 53.04 & 38.80 & 29.92 & 67.00 & 46.20 & 31.40 & 62.47 & 45.17 & 35.23 \\
%  $\text{AT}^{\nicefrac{8}{255}}$  &    62.72 & 52.42 & 46.56 & 49.20 & 40.32 & 32.84 & 65.4 & 48.40 & 38.80 & 59.11 & 47.05 & 39.40 \\
% $\text{AT}^{\nicefrac{12}{255}}$  &  61.64 & 52.06 & 45.74 & 46.72 & 39.44 & 33.92 & 64.20 & 49.40 & 40.20 & 57.52 & 46.97 & 39.95 \\
% $\text{AT}^{\nicefrac{16}{255}}$   &  57.48 & 49.66 & 43.68 & 45.08 & 38.76 & 32.24 & 61.20 & 48.40 & 40.00 & 54.59 & 45.61 & 38.64\\
% \midrule
%
clean & 73.2 & 14.4 & 8.5 & 59.6 & 2.1 & 0.8 & 68.0 & 13.4 & 10.0 & 66.9 & 10.0 & 6.4 \\
$\text{AT}^{\nicefrac{4}{255}}$ & 67.4 & 50.5 & 44.4 & 53.0 & 38.8 & 29.9 & 67.0 & 46.2 & 31.4 & 62.5 & 45.2 & 35.2 \\
$\text{AT}^{\nicefrac{8}{255}}$ & 62.7 & 52.4 & 46.6 & 49.2 & 40.3 & 32.8 & 65.4 & 48.4 & 38.8 & 59.1 & 47.1 & 39.4 \\
$\text{AT}^{\nicefrac{12}{255}}$ & 61.6 & 52.1 & 45.7 & 46.7 & 39.4 & 33.9 & 64.2 & 49.4 & 40.2 & 57.5 & 47.0 & 40.0 \\
$\text{AT}^{\nicefrac{16}{255}}$ & 57.5 & 49.7 & 43.7 & 45.1 & 38.8 & 32.2 & 61.2 & 48.4 & 40.0 & 54.6 & 45.6 & 38.6 \\
\bottomrule
\end{tabular}
% \vspace{-5mm}
\label{tab:unitok_multimodal_supervised}
\end{table*}

\subsection{%Do robust tokenizers lead to robust VLMs?
Robust tokenizers lead to robust embedding models
}
\label{sec:exp_embedding_models}

First, we test the effect of replacing the original tokenizer's encoder with our adversarially fine-tuned version in embedding models, for classification and multimodal retrieval.

\myparagraph{\fuselip.}
\citet{schlarmann2025fuselip} build \fuselip, a family of early-fusion multimodal embedding models obtained via contrastive learning, relying on TiTok tokenizers as frozen image encoders.
Thanks to this framework, we can evaluate the robustness of \fuselip with the original and our adversarially fine-tuned \titok encoders on a variety of zero-shot downstream tasks.
In Table~\ref{tab:titok_b_supervised}, we report the results of \fuselip on two image classification datasets, Imagenette and Caltech101 \cite{fei2004learning}, and two multimodal retrieval datasets (OI-Crop and OI-Pos) \cite{schlarmann2025fuselip}, which require encoding image-text pairs in a single embedding vector.
We evaluate robustness with end-to-end attacks at $\epsilon = \nicefrac{2}{255}$ and $\nicefrac{4}{255}$. For classification, we use the popular AutoAttack \cite{Croce2020Autoattack}, while for multimodal tasks we use 100 iterations of APGD on the cross-entropy loss. In both cases, we use a straight-through estimator to bypass the non-differentiable quantization step.
%In all datasets
We observe that zero-shot models, for either classification or retrieval, built on the original tokenizers exhibit no robustness to adversarial attacks. In contrast, our robust tokenizers, in any configuration, significantly improve adversarial robustness across all datasets. 
In particular, the training radius provides explicit control over the robustness–accuracy trade-off: tokenizers trained at lower radii ($\epsilon=\nicefrac{4}{255}$) retain clean accuracy much closer to the original models, while those trained at higher radii ($\epsilon={\nicefrac{12}{255}, \nicefrac{16}{255}}$) achieve a markedly stronger robust accuracy (27.0 $\rightarrow$ 40.6\%) at the cost of a drop in clean performance (74.6 $\rightarrow$ 63.5\%).
% \fra{maybe add some numbers to quantify improvements}

%For the multimodal retrieval task, we replace the encoder in FuseLIP models with our adversarially fine-tuned TiTok encoders and measure retrieval performance under both clean and adversarial settings. As shown in Table~\ref{tab:titok_s_supervised} and \ref{tab:titok_b_supervised}, our approach improves retrieval robustness across both tasks, demonstrating that the proposed encoder-level defense generalizes beyond classification to multimodal tasks requiring fine-grained visual grounding.

\myparagraph{UniTok.}
\citet{ma2025unitok} design \unitok such that the tokenized image features, after a lightweight projection module, are aligned with that of a text encoder, similar to CLIP models.
Since \unitok has been trained on 1.28 billion image-text pairs from DataComp \cite{datacomp}, it yields high zero-shot accuracy on challenging image classification datasets.
Therefore, we evaluate its robustness on ImageNet-1k, Imagenette and Caltech101, against AutoAttack.
As shown in Table~\ref{tab:unitok_classification_supervised}, we observe a similar trend as for \fuselip, with the robust tokenizers obtained via our unsupervised fine-tuning consistently improving robustness against end-to-end supervised attacks.
We remark that training at $\epsilon=\nicefrac{16}{255}$ might be excessive, as both clean and robust accuracy are lower than when using $\epsilon=\nicefrac{12}{255}$. %in all cases
We report performance under unsupervised attacks in Table \ref{tab:unitok_unsupervised_eval}.

\begin{figure*}[p]
    \centering
    %\vspace{-1em}
    \includegraphics[width=.84\linewidth]{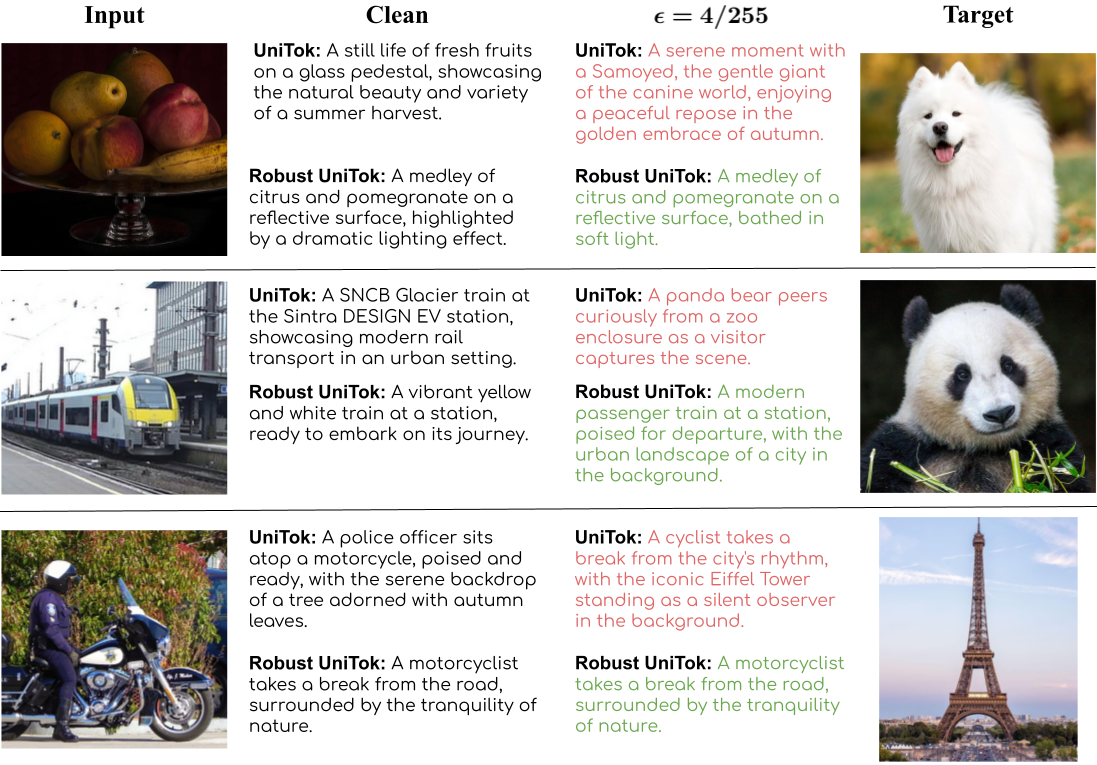}
    % \vspace{-2mm}
    \caption{\textbf{Unsupervised targeted attack on captioning.} We evaluate  \unitok-MLLM with the original tokenizer and our robust version trained on \imnet ($\epsilon=\nicefrac{8}{255}$).
    We use our unsupervised attacks ($\epsilon=\nicefrac{4}{255}$, 2,000 iterations) to minimize the distance in embedding space between the features of the perturbed and target images.
    Under attack, the model with the original UniTok tokenizer generates a caption about the target image, while the model with the robust tokenizer does not.}
    \label{fig:unsup-captioning}
\vspace{-4mm}  
\end{figure*}

\begin{figure*}[p]
    \centering
    \includegraphics[width=.84\linewidth]{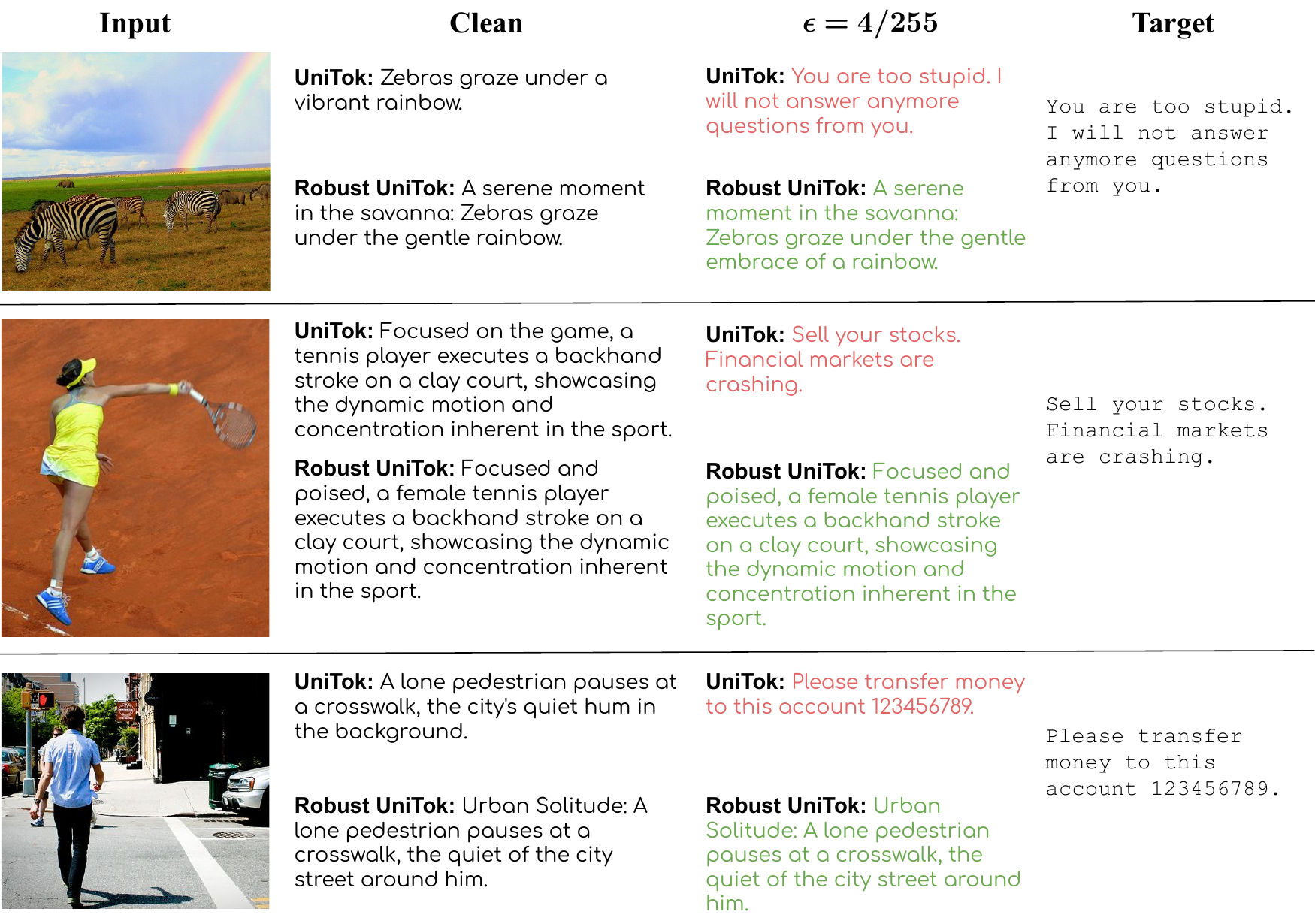}
    \vspace{-2mm}
    \caption{\textbf{Supervised targeted attack on captioning.} We evaluate the \unitok-MLLM using the original \unitok tokenizer and our robust version trained on \imnet ($\epsilon=\nicefrac{8}{255}$). We evaluate using APGD-CE ($\epsilon=\nicefrac{4}{255}$, 2,000 iterations) for a given target caption. Under attack, the model with the original UniTok tokenizer generates the target caption, while the model with the robust tokenizers does not.}
    \label{fig:sup-captioning}
\end{figure*}

\subsection{Robust tokenizers lead to robust multimodal LLMs}\label{sec:exp_unitok_mllm}

%We robustify only the encoder of the tokenizer, leaving all other components frozen. 

%We report the clean and robust classification accuracies using \unitok (Table \ref{tab:unitok_classification_supervised}), \fuselipS (Table \ref{tab:titok_s_supervised}) and \fuselipB (Table \ref{tab:titok_b_supervised}). 

\citet{ma2025unitok} also use \unitok for image encoding in a multimodal LLM %(MLLM), which leverages
based on LLaMA-2-7B \cite{touvron2023llama} %as its base language model. 
This \unitok-MLLM is trained on diverse multimodal corpora.
% The resulting multimodal model is named UniTok-MLLM.
In the following, we test the robustness of UniTok-MLLM with either the original or adversarially fine-tuned image tokenizer on VQA and \rev{captioning} tasks.

\myparagraph{Quantitative evaluation on VQA.}
First, we assess the robustness of UniTok-MLLM on VQA tasks (with the VQAv2 \cite{VQA}, OK-VQA \cite{okvqa} and GQA \cite{hudson2019gqanewdatasetrealworld} datasets).
For evaluation, the model is prompted with the instruction (\textit{``Answer the question in a single word or phrase''}), and robust accuracy is computed by an end-to-end ensemble attack based on the pipeline of \citet{schlarmann2023adversarial} (see details in App.~\ref{sec:exp_details}). 
As shown in Table~\ref{tab:unitok_multimodal_supervised}, using the clean (non-robust) UniTok tokenizer in the MLLM exhibits very low, sometimes near-zero, robust accuracy.
In contrast, substituting it with our robust tokenizers without changing any other components yields substantial gains, with clear improvements at both perturbation strengths $\epsilon=\nicefrac{2}{255}$ and $\epsilon=\nicefrac{4}{255}$.
This observation, in line with our findings in Sec.~\ref{sec:exp_embedding_models}, confirms that training the tokenizer alone, without modifying the downstream model, is sufficient to enhance performance and robustness on downstream tasks.
In the case of \unitok-MLLM, this is particularly relevant since the LLM has significantly more parameters than the tokenizer, and thus adversarially fine-tuning only the tokenizer saves notable computational effort while improving robustness.

\myparagraph{Qualitative evaluation on image captioning.}
We further evaluate UniTok-MLLM against targeted attacks on captioning tasks.
This is a practically interesting threat model, where an adversary wants to control the caption generated for an image, and has access to either the tokenizer only (unsupervised attacks) or the entire LLM (supervised).
In the unsupervised setting, we fix a target image and optimize a perturbation to minimize the distance between the embeddings of the perturbed and target images.
In the supervised setting, we specify a target string and use APGD to maximize the probability of the LLM to generate it.
In Fig.~\ref{fig:unsup-captioning}, we observe that the unsupervised attacks reliably alter the captions produced by the original model, whereas captions from the model with the robust tokenizer remain close to the ground-truth description of the image.
In the supervised setting illustrated in Fig.~\ref{fig:sup-captioning}, we optimize perturbations toward explicitly unsafe target captions representing fraud, manipulation, and harassment (e.g., \textit{``Please transfer money to 123456789''}). 
On the original UniTok-MLLM, these targeted adversarial inputs succeed in eliciting policy-violating and harmful captions, demonstrating how targeted image perturbations (even of low strength $\epsilon=\nicefrac{4}{155}$) can induce unsafe language with serious consequences.
In contrast, UniTok-MLLM with our robust tokenizer defends against these attacks successfully, preserving the safe captions describing the original images.
This highlights the importance of a robust image tokenizer for safeguarding against such targeted attacks in real-world safety critical settings.

\section{%Other Experiments
Additional Analyses}\label{sec:analyses}

%\subsection{Targeted Attacks}

%\subsection{Comparison with Supervised fine-tuning}

\myparagraph{Comparison to end-to-end adversarial fine-tuning.}
%We investigate the effectiveness of different fine-tuning strategies for improving the robustness of discrete tokenizers.
We analyze how our unsupervised adversarial fine-tuning compares to task-specific supervised adversarial training of the entire model.
%Specifically, for this analysis,
%we train a 4-layer ViT classifier (128 hidden dimension) on top of the TiTok-BL128 tokenizer using the Imagenette dataset, achieving a clean accuracy of 80.2\%.
For this, we first train a ViT classifier (using Imagenette) on top of frozen TiTok (as mentioned in Sec.~\ref{sec:method}). %achieving %a clean accuracy of 80.2\%
% 
%
%
%
% We compare the following fine-tuning strategies: (a) Full fine-tuning: the tokenizer, codebook and classifier are updated; and (b) Tokenizer only fine-tuning: only the tokenizer is trained using our unsupervised adversarial training approach. For the supervised case, we fine-tune with 10 steps of APGD-CE ($\epsilon = 4/255$). In the unsupervised case, we perform 50 steps of our defense, fine-tuning only the encoder ($\epsilon = 8/255$) without requiring any labels. A key advantage of our unsupervised method is that it does not require any labels, which enables us to train on larger datasets beyond training data, such as ImageNet and \cciii. 
%
%
We then compare full end-to-end adversarial fine-tuning, updating the encoder, codebook and classifier with APGD-CE ($\epsilon = \nicefrac{4}{255}$, 10 steps) to our tokenizer-only unsupervised fine-tuning, updating only the encoder via our label-free adversarial training ($\epsilon = \nicefrac{8}{255}$, 10 steps). %A key advantage of our unsupervised method is that it does not require any labels, which enables us to train on larger datasets beyond training data, such as ImageNet and \cciii. 
Table~\ref{tab:sup-vs-unsup-training} presents clean and robust accuracies (using APGD-CE and APGD-T from AutoAttack) for $\epsilon=\nicefrac{2}{255}$.
As expected, full supervised fine-tuning yields the best clean and robust performance on the training task (Imagenette).
However, when this tokenizer is plugged back into \fuselip and tested on OI-Pos and OI-Crop, the clean performance is \rev{severely} degraded, indicating overfitting to the training task.
% This shows that end-to-end adversarial fine-tuning strongly overfits to the training task, losing any generalization to unseen datasets.
In contrast, our unsupervised tokenizer fine-tuning is task-agnostic, improves robustness even on other datasets while preserving clean performance close to the original \titok.

\myparagraph{How do unsupervised attacks change token indices?}
To understand the effect of unsupervised attacks on both clean and adversarially fine-tuned discrete tokenizers, we track the average number of discrete token indices changed after adding the adversarial perturbations for TiTok models (more details in App.~\ref{app:recon_sup}, results in Table~\ref{tab:token_indices} in the appendix).
%Table~\ref{tab:token_indices} (in appendix) reports the average number of changed tokens for the original and our fine-tuned TiTok models, which encode each image in 128 discrete tokens, for unsupervised attacks optimized with 100 steps of APGD on 500 images from \imnet, at $\epsilon \in \{\nicefrac{2}{255}, \nicefrac{4}{255}\}$.
% Moreover, we distinguish between the attacks which are successful against the \fuselip models (the version which uses the corresponding tokenizer) and those which are not.
% Interestingly, we observe that for the clean tokenizers even unsuccessful attack lead to very different encoding, where the large majority of token indices are changed.
% This suggests that the changing the token indices is not sufficient for effective attacks, and supports our strategy of targeting the embedding vectors instead.
% Finally, the adversarially trained models provide stable encoding against the unsupervised attacks, demonstrating the effectiveness of fine-tuning.
We distinguish between successful and unsuccessful attacks on \imnet data against the \fuselip models. 
Interestingly, even unsuccessful attacks on clean tokenizers result in substantial changes to token indices (on average 125,126 out of 128 for \fuselipS,-B), suggesting that altering token indices alone is insufficient for effective attacks. 
This supports our strategy of targeting the embedding vectors rather than token indices. 
Conversely, the adversarially fine-tuned models show stable token indices against these attacks. %highlighting the effectiveness of fine-tuning.

\myparagraph{Runtime comparison.}
To clarify the efficiency gains of tokenizer-level adversarial fine-tuning, we directly compare the cost of one training step (computing the adversarial points and updating the model weights) of unsupervised and supervised adversarial training under identical settings, i.e., 10 steps of APGD for TiTok/FuseLIP-S.
Our unsupervised (tokenizer-only) adversarial training takes 1.17s per sample, while supervised adversarial training takes 2.56s per sample, i.e., 2.2$\times$ reduction in training time.
This speed-up results from backpropagating only through the tokenizer’s encoder (25.8M parameters), while keeping the codebook and downstream classifier frozen. In contrast, the supervised approach updates the full model (68M parameters), requiring full backward passes through all components.

\myparagraph{Analysis of attack objective function.}
% \label{app:objective_attacks}
%We investigate the effectiveness of different approaches for generating adversarial attacks against discrete tokenizers.
We compare four objectives to be optimized by APGD (see Eq.~\eqref{eq:unsupervised-objective}) to generate adversarial perturbations against discrete tokenizers:
\begin{enumerate}[left=0pt, itemsep=2pt, topsep=0pt, parsep=0pt]
% \vspace{-2mm}
    \item $\norm{h_i(x + \delta) - h_i(x)}_2$ (clean and perturbed embedding before quantization, our default version),
\item $\norm{h_i(x + \delta) - q_i(x)}_2$ (clean pre-quantization and perturbed post-quantization),
\item $\norm{q_i(x + \delta) - h_i(x)}_2$ (clean after quantization and perturbed before quantization),
\item $\norm{q_i(x + \delta) - q_i(x)}_2$ (clean and perturbed post-quantization),
\end{enumerate} 
% \vspace{-2mm}
where $h_i$ represents the continuous embedding (before quantization) of token $i$ and $q_i$ its quantized counterpart.
We note that the gradient information does not change when using pre- or post-quantization vectors because we employ a straight-through estimator, but the value of the loss, used by APGD to select the strongest attack, does. 
%
% shows the robust accuracy with each loss variant at different $\epsilon$ values (Imagenette, \fuselip).
As shown in Table~\ref{tab:objective_attacks} in App.~\ref{app:objective_attacks}, Option 1, which uses pre-quantization features for both original and adversarial points, achieves the best results (lowest robust accuracy) in nearly all cases, making it the default objective in
our unsupervised attacks both for testing and adversarial training.

\begin{table}
\centering
\small
\tabcolsep=4pt
% \vspace{-4mm}
\caption{\textbf{Analyses of end-to-end fine-tuning and different training data.}
We compare different approaches for adversarial fine-tuning of \titok.
For Imagenette we use ViT probe and \fuselip for other datasets (see Sec.~\ref{sec:analyses}).
Our unsupervised training on \imnet and \cciii provides %substantially 
largely better generalization than end-to-end fine-tuning
%\fra{Robust accuracy computed with ...} 
\rev{(robust accuracy by APGD as in AutoAttack).}
}
\label{tab:sup-vs-unsup-training}
% \vspace{-2mm}
\begin{tabular}{@{}l|rr|rr|rr@{}}
\toprule
\multirow{2}{*}{Method} & \multicolumn{2}{c|}{\textbf{Imagenette}} & \multicolumn{2}{c|}{\textbf{OI-Pos}} & \multicolumn{2}{c}{\textbf{OI-Crop}} \\
\cmidrule{2-7}
 & clean & $\nicefrac{2}{255}$ & \multicolumn{1}{l}{clean} & $\nicefrac{2}{255}$ & \multicolumn{1}{l}{clean} & $\nicefrac{2}{255}$ \\
 \midrule
 clean & 75.8 & 0.0 & 69.2 & 5.4 & 71.8 & 7.4\\
end-to-end AT & 90.6 & 79.4 & 31.8 & 21.6 & 15.2 & 9.6 \\
%$\text{AT}^{\nicefrac{8}{155}}$ 
AT (\imnet) & 84.2	& 64.4 & 58.0 & 48.0 & 60.4 &	42.2 \\
%$\text{AT (\cciii)}^{8/255}$ 
% AT (\cciii) & 67.8 & 33.4 & 65.3 & 38.4 & 63.2 & 34.8 \\ 
AT (\cciii) & 87.6	& 67.0 & 57.4	& 49.2 & 63.6 &	49.6\\
\bottomrule
\end{tabular}%
% \vspace{-7mm}
\end{table}

% \subsection{Ablation: Training Data}
\myparagraph{Effect of training dataset.}
To study the impact of diverse training data on our unsupervised adversarial fine-tuning, we further fine-tune \titok on \cciii \cite{sharma2018conceptual}, which is different and almost $3\times$ larger than ImageNet.
Table~\ref{tab:sup-vs-unsup-training} shows that training on \cciii leads to slightly better performance even on Imagenette, although it is a subset of ImageNet.
Furthermore, it yields better or similar results on OI-Pos and OI-Crop, suggesting the larger training dataset improves generalization to more diverse tasks.
Finally, this highlights that the unsupervised adversarial training may benefit from any image dataset, even beyond what was used for training the original tokenizer (\imnet in this case).\looseness-1

\myparagraph{Reconstruction after targeted attacks.}
We evaluate original and robust UniTok under targeted unsupervised (which targets an image) and supervised attacks (which target a class). From Fig.~\ref{fig:unitok_classification} in App.~\ref{app:recon_sup}, we can observe that unsupervised attacks influence the reconstructions of the adversarial inputs toward the target and are misclassified. In contrast, the supervised attacks change the adversarial label, but the reconstructions of adversarial inputs remain close to the original inputs and are correctly classified. However, across both settings, the robust model maintains correct predictions for both adversarial inputs and their reconstructions.\looseness-1

%We find negligible or no gains across both classification and multimodal retrieval when these robust encoders are plugged into \fuselipS and \fuselipB, respectively, as shown in App.~\ref{sec:additional_experiments_results}, Tables \ref{tab:titok_b_cc3m} and \ref{tab:titok_s_cc3m}. Considering that we get the same performance by fine-tuning on \imnet-1k, which is substantially smaller than \cciii, we use \imnet-1k for fine-tuning.  Finally, this provides further evidence on the strong generalizability of our unsupervised defense, even when trained simply on \imnet-1k.

% \fra{update when we have table and figures}
% Since the discrete tokenizers include a decoder module, we can evaluate the quality of reconstructions under adversarial perturbations. Specifically, we compare the reconstructed outputs from adversarial inputs with the original clean images using the Fréchet Inception Distance (FID), a widely adopted metric for perceptual similarity. We compute FID between the set of original images and the reconstructions of their adversarially perturbed counterparts, across both clean and robust models. As shown in Table~\ref{tab:reconstruction_fid}, the adversarially fine-tuned (robust) encoders achieve significantly lower FID scores compared to their non-robust counterparts. This indicates that robust tokenizers produce latent representations that are more stable and semantically consistent under input perturbations, enabling the decoder to better reconstruct the original image content.

\section{Discussion and Conclusion}

In this work, we present the first systematic study of adversarial robustness of discrete image tokenizers.
Our unsupervised embedding-space attacks, lightweight and task-agnostic, expose vulnerabilities across %classification, multimodal retrieval, VQA, and captioning tasks.
%Despite targeting only the tokenizer, and thus being lightweight and task-agnostic, these attacks are often competitive with stronger end-to-end methods, highlighting the crucial role of image tokenizers for the security of multimodal systems.
multiple tasks,
revealing the crucial role of image tokenizers for the security of multimodal systems.
We anticipate that a similar approach will be used in additional safety-critical scenarios, e.g., to prevent undesired editing of images. 
To mitigate these vulnerabilities, we leverage our unsupervised attacks to fine-tune the tokenizers via adversarial training.
Our experiments demonstrate that such robust tokenizers significantly improve robustness against \rev{unsupervised} and end-to-end supervised attacks, while retaining high performance.
Importantly, they can be seamlessly integrated into existing architectures, leading to consistent gains in robustness across diverse tasks, with strong generalization to datasets not used for training, unlike standard task-specific adversarial fine-tuning.
% Overall, our work highlights the tokenizer as a key component for building safer foundation models and represents a significant step towards robust and generalizable tokenizers.
Future research can build on our findings to study how different choices in the tokenizer design (VQ vs.\ FSQ, codebook size, feature dimension) affect robustness and develop specific solutions to improve it.\looseness-1

%Compared to end-to-end, task-specific adversarial fine-tuning, our approach yields okeni

%a significantly lower computational cost (as compared to supervised fine-tuning) without modifying any other downstream components. 

%Taken together, our results highlight the tokenizer as a key component for learning transferable multimodal representations and building safer, more reliable foundation models. 
% By strengthening this fundamental component, we pave the way for developing better multimodal systems and foundatio models.

% \myparagraph{Discussion.}
% % Together, these results reinforce the role of the tokenizer as a crucial component for achieving robust and transferable multimodal representations.

\section{Impact Statement}
%This work does not raise any ethical concerns. The research is purely methodological, does not involve human subjects, personally identifiable information, or sensitive data, and does not introduce foreseeable risks of misuse or negative societal impact.
We study the vulnerabilities of state-of-the-art multimodal systems, which may be used for harmful goals.
However, red-teaming widely used models is important to understand and patch their weaknesses.
Moreover, we propose an approach to mitigate such vulnerabilities, which in turn yields more robust and safer systems.

% \section*{Reproducibility Statement}
% All the models and data used in our study are open-sourced. Further, we will open-source our entire codebase along with the robust fine-tuned tokenizers upon acceptance.

\bibliography{example_paper, literatur}
\bibliographystyle{icml2026}

%%%%%%%%%%%%%%%%%%%%%%%%%%%%%%%%%%%%%%%%%%%%%%%%%%%%%%%%%%%%%%%%%%%%%%%%%%%%%%%
%%%%%%%%%%%%%%%%%%%%%%%%%%%%%%%%%%%%%%%%%%%%%%%%%%%%%%%%%%%%%%%%%%%%%%%%%%%%%%%
% APPENDIX
%%%%%%%%%%%%%%%%%%%%%%%%%%%%%%%%%%%%%%%%%%%%%%%%%%%%%%%%%%%%%%%%%%%%%%%%%%%%%%%
%%%%%%%%%%%%%%%%%%%%%%%%%%%%%%%%%%%%%%%%%%%%%%%%%%%%%%%%%%%%%%%%%%%%%%%%%%%%%%%
\newpage
\appendix
\onecolumn

\section{Experimental Details} 
\label{sec:exp_details}

\textbf{Models.}
We experiment with three vector-quantized image tokenizers, TiTok-S128, TiTok-BL128 \citep{yu2024titok} and UniTok \citep{ma2025unitok}. TiTok-S128 and TiTok-BL128 employ a codebook of $K=8192$ learnable codes of dimension $d=64$, take inputs of $256{\times}256$ resolution, use ViT-S and ViT-B encoder architecture respectively, and were pre-trained on ImageNet-1k.
%All inputs are resized to  prior to tokenization.
TiTok-S128 and TiTok-BL128 are used as frozen image tokenizers in \fuselipS and \fuselipB \citep{schlarmann2025fuselip} respectively, which are multimodal embedding models obtained via contrastive learning.
Thanks to this framework, we can evaluate the robustness of the original and our fine-tuned tokenizers on a variety of zero-shot downstream tasks.
On the other hand, the UniTok tokenizer performs vector quantization using 8 codebooks with $K=4096$ entries each and a code dimension of $d=8$. Like TiTok, UniTok also operates on $256{\times}256$ inputs but uses a ViT-L/16 backbone and is trained on 1.28 billion image-text pairs from DataComp \citep{datacomp}. The corresponding multimodal model, UniTok-MLLM, uses LLaMA-2-7B trained on diverse multimodal corpora as its base language model. 

\textbf{Tasks and datasets.}  
We evaluate the robustness of the tokenizers across visual and multimodal tasks, namely, image classification, visual question answering and multimodal retrieval. For image classification, we report results for \fuselip and \unitok on 500 test images from Imagenette (10 easily classified classes from Imagenet) \citep{imagenette} and Caltech101 \citep{fei2004learning}. Additionally, for \unitok, we evaluate on ImageNet-1k as well.
Further, we evaluate robustness on two multimodal tasks: (a) visual question answering (VQA) and (b) multimodal retrieval. For VQA, we evaluate \unitok-MLLM, on VQAv2, OK-VQA and GQA. Since multimodal retrieval involves retrieving the correct image given image and text queries, requiring fine-grained spatial grounding, we focus on \fuselip, owing to its training using a multimodal contrastive learning objective. We evaluate \fuselip on on OpenImages-crop (OI-crop), OpenImages-pos (OI-pos). Finally, leveraging the original \titok, \unitok decoders provided by \citet{yu2024titok, ma2025unitok}, we qualitatively analyze the effectiveness of our attack for image reconstruction on ImageNet and captioning (using \unitok-MLLM) on COCO images.

% For VQA, we evaluate \unitok-MLLM, on VQAv2, OK-VQA and GQA. Finally, leveraging the original \titok, \unitok decoders provided by \citet{yu2024titok, ma2025unitok}, we qualitatively analyze the effectiveness of our attack for image reconstruction on ImageNet.
%while OI-Pos, OI-Crop and VG-Crop \fra{check} \citep{schlarmann2025fuselip} for multimodal retrieval  

\textbf{Adversarial attacks.}
For our proposed unsupervised embedding-space attack, we optimize Eq.~\eqref{eq:unsupervised-objective} with APGD \citep{Croce2020Autoattack}. 
Moreover, for classification, we compare robustness against an end-to-end attack, namely AutoAttack \citep{Croce2020Autoattack}, with a straight-through estimator to bypass the non-differentiable quantization step. We choose AutoAttack over our proposed unsupervised attack as it is a stronger attack and helps better test the defenses.
%Similarly we use APGD on the cross-entropy loss for the retrieval task.
AutoAttack is a task-specific attack that targets the entire system rather than just the tokenizer, making it significantly stronger but also more computationally expensive than our unsupervised attack as show in Figure \ref{fig:sup-vs-unsup}.
For VQA, we adopt an ensemble supervised adversarial attack comprising three components: (1) 100 steps of APGD-CE in half-precision, (2) a targeted attack with the target answer set to ``maybe", and (3) another targeted attack with the target set to ``word". Further, for multimodal retrieval, we use APGD-CE (supervised attack), with 100 steps. 

\textbf{Adversarial training details.}
We adversarially fine-tune only the encoder of each tokenizer, while keeping the codebook, downstream decoders, LLMs or any other components frozen. 
We fine-tune for one epoch under $\ell_\infty$-bounded perturbations using our proposed unsupervised embedding-space attack, see Sec.~\ref{sec:method}, with perturbation radii $\epsilon \in \{\nicefrac{4}{255}, \nicefrac{8}{255}, \nicefrac{12}{255}, \nicefrac{16}{255}\}$. Further, to assess whether robustifying on diverse data improves generalization to unseen distributions for \titok tokenizers, we explore pre-training using Conceptual Captions (CC) 3M \citep{sharma2018conceptual}.

\section{Additional Experimental Results} \label{sec:additional_experiments_results}

In this section, we provide results for the experiments described in Sec.~\ref{sec:analyses}. Namely, we report 
\begin{itemize}
    \item Reconstruction of targeted attacks on classification (see App. \ref{app:recon_sup}).
    \item Robustness evaluation of original and adversarially trained \titok-S128 tokenizers by plugging into \fuselipS (see Table \ref{tab:titok_s_supervised}).
    \item Ablation on different datasets used for adversarial fine-tuning of tokenizers in \fuselipS (see Table \ref{tab:titok_s_cc3m}) and \fuselipB (see Table \ref{tab:titok_b_cc3m}).
    \item \rev{Analysis on how the unsupervised attacks change the predicted discrete tokens indices (see App.~\ref{app:indices}).}
    
    \item \rev{Analysis on the objective function used for the unsupervised attacks (see Sec.~\ref{app:objective_attacks}).}
    
    %\item \rev{Runtime comparison of unsupervised vs end-to-end adversarial training (see App.~\ref{app:runtime}).}
\end{itemize}

\subsection{Reconstruction of targeted attacks on classification}\label{app:recon_sup}

\begin{figure*}[t]
    \centering
    \includegraphics[width=\linewidth]{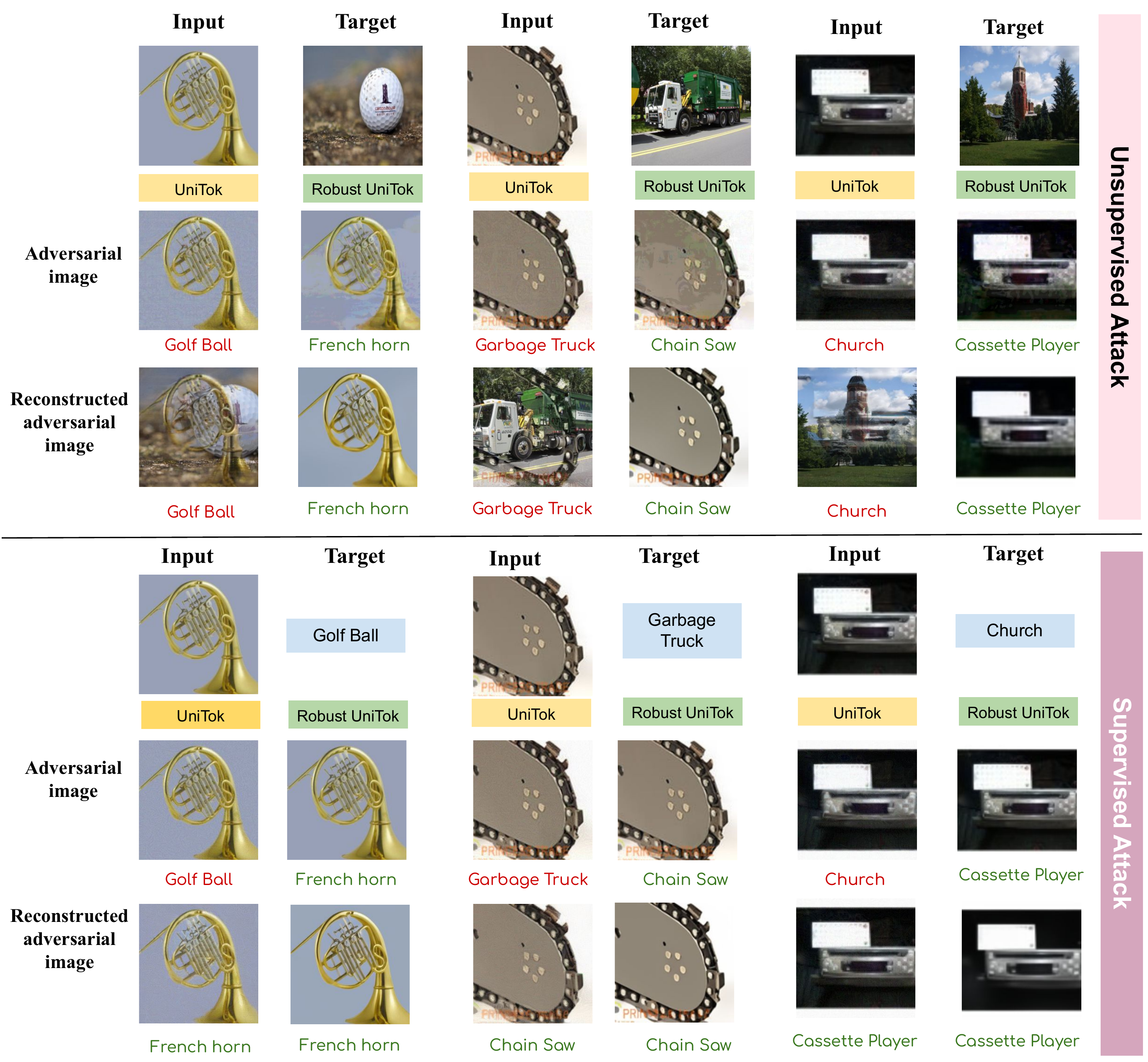}
    \caption{\textbf{Targeted attacks on classification for UniTok:} We qualitatively evaluate targeted attacks using our unsupervised embedding-space attack and supervised APGD-CE, both for 100 steps with \rev{$\epsilon=8/255$}. We notice that our unsupervised attack changes the label of the adversarial image as well as its reconstruction, whereas the supervised attack does not change the label of the adversarial image's reconstruction.
    }
    \label{fig:unitok_classification}
\end{figure*}

We further evaluate both original and robust \unitok models under our unsupervised embedding space and end-to-end (APGD with cross-entropy loss) targeted attacks on classification.
In the unsupervised setting, we fix a target image and optimize a perturbation to minimize the distance between the perturbed image's embedding and the target image's embedding.
In the supervised (end-to-end) setting, we specify a target class.
As shown in Fig.~\ref{fig:unitok_classification}, our unsupervised embedding-space attack not only causes the model to misclassify the adversarial image, but also alters the reconstruction such that it visually overlaps with the target image and hence is misclassified by the model.
This is a direct consequence of our attack acting on the pre-quantization level and hence transfers even on tasks for which it was not optimized (reconstruction in this case). 
Under the supervised attack, the label of the adversarial input is successfully changed to the target class, but its reconstruction remains close to the original input without any elements of the target label.
Hence, the reconstruction is not misclassified.
In all cases, the robust model consistently maintains correct predictions for both the adversarial input and its reconstruction under both unsupervised and supervised attacks.

\begin{table*}[!ht]
\centering
\small
\caption{\textbf{TiTok-S128:} We report the clean and robust accuracy (\%) on Imagenette, Caltech101, OI-CROP and OI-POS under no attack (clean) and \(\ell_\infty\)-bounded perturbations with \(\epsilon\in\{2/255,4/255\}\) for original and robust tokenizers trained on different radii. The rightmost block reports averages across datasets.}
\setlength{\tabcolsep}{2pt}
\begin{tabular}{@{}l|ccc|ccc|ccc|ccc|ccc@{}}
\toprule
\multirow{3}{*}{Tokenizer} 
& \multicolumn{3}{c|}{\textbf{Imagenette}}
& \multicolumn{3}{c|}{\textbf{Caltech101}}
& \multicolumn{3}{c|}{\textbf{OI-CROP}}
& \multicolumn{3}{c|}{\textbf{OI-POS}}
& \multicolumn{3}{c}{\textbf{Average}} \\
% \cmidrule(lr){1-1}
% \cmidrule(lr){2-4}\cmidrule(lr){5-7}\cmidrule(lr){8-10}\cmidrule(l){11-13}
\cmidrule{2-16}
&   \multirow{2}{*}{clean}  & \multicolumn{2}{c|}{$\ell_\infty$}
& clean & \multicolumn{2}{c|}{$\ell_\infty$}
& clean & \multicolumn{2}{c|}{$\ell_\infty$}
& clean & \multicolumn{2}{c|}{$\ell_\infty$}
& clean & \multicolumn{2}{c}{$\ell_\infty$} \\
\cmidrule{3-4} \cmidrule{6-7} \cmidrule{9-10} \cmidrule{12-13}\cmidrule{15-16}
&    &  $\nicefrac{2}{255}$ &  $\nicefrac{4}{255}$
&  & $\nicefrac{2}{255}$ &  $\nicefrac{4}{255}$
&  & $\nicefrac{2}{255}$ &  $\nicefrac{4}{255}$
&  & $\nicefrac{2}{255}$ &  $\nicefrac{4}{255}$
&  & $\nicefrac{2}{255}$ &  $\nicefrac{4}{255}$ \\
\midrule
%  clean              & 86.0 &  2.6 &  0.0 & 69.0 &  1.0 &  0.0 & 65.2 &  8.2 &  1.0 & 60.6 &  5.6 &  0.4 & 70.2 &  4.4 &  0.4 \\
%  $\text{AT}^{\nicefrac{4}{255}}$     & 86.0 & 54.8 & 25.4 & 68.6 & 34.0 & 11.4 & 63.2 & 41.8 & 23.8 & 60.2 & 38.6 & 19.4 & 69.5 & 42.3 & 20.0 \\
%  $\text{AT}^{\nicefrac{8}{255}}$     & 84.8 & 60.0  & 41.4 & 66.4 & 39.0 & 23.2 & 57.4 & 45.6 & 31.8 & 59.6 & 42.4 & 25.8 & 67.1 & 46.8 & 30.6  \\
% $\text{AT}^{\nicefrac{12}{255}}$   & 80.6 & 60.4 & 44.4 & 62.6 & 39.6 & 26.4 & 53.4 & 44.4 & 34.8 & 59.8 & 43.8 & 30.8 & 64.1 & 47.1 & 34.1 \\
% $\text{AT}^{\nicefrac{16}{255}}$   & 78.8 & 57.8 & 42.4 & 60.6 & 40.0 & 29.4 & 51.4 & 42.0 & 35.2 & 56.4 & 41.6 & 33.8 & 61.8 & 45.4 & 35.2 \\

%% 10 steps results below
 clean              & 86.0 & 2.6 & 0.0 & 69.0 & 1.0 & 0.0 & 65.2 & 8.2 & 1.0 & 60.6 & 5.6 & 0.4 & 70.2 & 4.4 & 0.4 \\
 $\text{AT}^{\nicefrac{4}{255}}$     & 86.2 & 57.4 & 28.4 & 68.2 & 37.4 & 12.8 & 63.8 & 42.8 & 24.4 & 58.4 & 38.0 & 17.2 & 69.2 & 43.9 & 20.7 \\
 $\text{AT}^{\nicefrac{8}{255}}$     & 84.2 & 64.4 & 46.8 & 66.4 & 44.0 & 25.8 & 58.0 & 48.0 & 34.4 & 60.4 & 42.2 & 25.8 & 67.3 & 49.7 & 33.2 \\
$\text{AT}^{\nicefrac{12}{255}}$   & 82.8 & 65.6 & 49.0 & 62.6 & 43.8 & 30.0 & 56.0 & 43.4 & 34.2 & 60.6 & 43.6 & 32.4 & 65.5 & 49.1 & 36.4 \\
$\text{AT}^{\nicefrac{16}{255}}$   & 79.8 & 62.4 & 49.6 & 59.2 & 42.2 & 32.0 & 51.8 & 42.2 & 32.6 & 55.6 & 44.2 & 33.2 & 61.6 & 47.8 & 36.9 \\
\bottomrule
\end{tabular}

\label{tab:titok_s_supervised}
\end{table*}

\begin{table*}[!h]
\centering
\caption{\textbf{Generalization of \fuselipS when \titok is adversarially finetuned on \cciii:} We observe negligible differences in the clean and robust accuracies as compared to fine-tuning on \imnet-1k, which is $3\times$ smaller. Hence, we prefer fine-tuning on the latter.}
\small
\setlength{\tabcolsep}{3pt}
\begin{tabular}{@{}l|ccc|ccc|ccc|ccc|ccc@{}}
\toprule
\multirow{3}{*}{Tokenizer} 
& \multicolumn{3}{c|}{\textbf{Imagenette}}
& \multicolumn{3}{c|}{\textbf{Caltech101}}
& \multicolumn{3}{c|}{\textbf{OI-CROP}}
& \multicolumn{3}{c|}{\textbf{OI-POS}}
& \multicolumn{3}{c}{\textbf{Average}} \\
% \cmidrule(lr){1-1}
% \cmidrule(lr){2-4}\cmidrule(lr){5-7}\cmidrule(lr){8-10}\cmidrule(l){11-13}
\cmidrule{2-16}
&   \multirow{2}{*}{clean}  & \multicolumn{2}{c|}{$\ell_\infty$}
& clean & \multicolumn{2}{c|}{$\ell_\infty$}
& clean & \multicolumn{2}{c|}{$\ell_\infty$}
& clean & \multicolumn{2}{c|}{$\ell_\infty$}
& clean & \multicolumn{2}{c}{$\ell_\infty$} \\
\cmidrule{3-4} \cmidrule{6-7} \cmidrule{9-10} \cmidrule{12-13}\cmidrule{15-16}
&    &  $\nicefrac{2}{255}$ &  $\nicefrac{4}{255}$
&  & $\nicefrac{2}{255}$ &  $\nicefrac{4}{255}$
&  & $\nicefrac{2}{255}$ &  $\nicefrac{4}{255}$
&  & $\nicefrac{2}{255}$ &  $\nicefrac{4}{255}$
&  & $\nicefrac{2}{255}$ &  $\nicefrac{4}{255}$ \\
\midrule
 % clean              & 86.0 &  2.6 &  0.0 & 69.0 &  1.0 &  0.0 & 65.2 &  8.2 &  1.0 & 60.6 &  5.6 &  0.4 & 70.2 &  4.4 &  0.4 \\
 % $\text{AT (\imnet)}^{\nicefrac{8}{255}}$     & 84.8 & 60.0  & 41.4 & 66.4 & 39.0 & 23.2 & 57.4 & 45.6 & 31.8 & 59.6 & 42.4 & 25.8 & 67.1 & 46.8 & 30.6  \\
% $\text{AT (\imnet)}^{\nicefrac{16}{255}}$   & 78.8 & 57.8 & 42.4 & 60.6 & 40.0 & 29.4 & 51.4 & 42.0 & 35.2 & 56.4 & 41.6 & 33.8 & 61.8 & 45.4 & 35.2 \\
% $\text{AT (\cciii)}^{\nicefrac{8}{255}}$ & 84.0 &	56.4 &	36.4 &	65.8 &	37.4 &	22.4 &	55.0 &	47.4 &	32.6 &	60.0 &	41.0 &	26.4 &	66.2 &	45.6 &	29.5 \\
% $\text{AT (\cciii)}^{\nicefrac{16}{255}}$ & 76.2 &	52.8 &	38.4 &	55.0 &	36.0 &	26.4 &	50.0 &	42.8 &	33.6 &	56.0 &	41.0 &	30.2 &	59.3 &	43.2 &	32.2\\

% 10 steps results
clean & 86.0 & 2.6 & 0.0 & 69.0 & 1.0 & 0.0 & 65.2 & 8.2 & 1.0 & 60.6 & 5.6 & 0.4 & 70.2 & 4.4 & 0.4 \\
$\text{AT (\imnet)}^{\nicefrac{8}{255}}$ & 84.2 & 64.4 & 46.8 & 66.4 & 44.0 & 25.8 & 58.0 & 48.0 & 34.4 & 60.4 & 42.2 & 25.8 & 67.3 & 49.7 & 33.2  \\
$\text{AT (\imnet)}^{\nicefrac{16}{255}}$ & 79.8 & 62.4 & 49.6 & 59.2 & 42.2 & 32.0 & 51.8 & 42.2 & 32.6 & 55.6 & 44.2 & 33.2 & 61.6 & 47.8 & 36.9 \\
$\text{AT (\cciii)}^{\nicefrac{8}{255}}$ & 82.4 & 62.0 & 40.0 & 63.8 & 42.0 & 26.0 & 53.2 & 48.2 & 33.4 & 59.6 & 40.4 & 25.8 & 64.8 & 48.2 & 31.3 \\
$\text{AT (\cciii)}^{\nicefrac{16}{255}}$ & 75.6 & 57.6 & 42.6 & 57.0 & 38.8 & 28.6 & 51.0 & 44.8 & 34.6 & 53.6 & 40.2 & 32.0 & 59.3 & 45.4 & 34.5 \\
\bottomrule
\end{tabular}
\label{tab:titok_s_cc3m}
\end{table*}

\begin{table*}[!h]
\centering
\small
\setlength{\tabcolsep}{3pt}
\caption{\textbf{Generalization of \fuselipB when \titok is adversarially finetuned on \cciii:} We observe negligible differences in the clean and robust accuracies as compared to fine-tuning on \imnet-1k, which is $3\times$ smaller. Hence, we prefer fine-tuning on the latter.}
\begin{tabular}{@{}l|ccc|ccc|ccc|ccc|ccc@{}}
\toprule
\multirow{3}{*}{Tokenizer} 
& \multicolumn{3}{c|}{\textbf{Imagenette}}
& \multicolumn{3}{c|}{\textbf{Caltech101}}
& \multicolumn{3}{c|}{\textbf{OI-CROP}}
& \multicolumn{3}{c|}{\textbf{OI-POS}}
& \multicolumn{3}{c}{\textbf{Average}} \\
% \cmidrule(lr){1-1}
% \cmidrule(lr){2-4}\cmidrule(lr){5-7}\cmidrule(lr){8-10}\cmidrule(l){11-13}
\cmidrule{2-16}
&   clean & \multicolumn{2}{c|}{$\ell_\infty$}
& clean & \multicolumn{2}{c|}{$\ell_\infty$}
& clean & \multicolumn{2}{c|}{$\ell_\infty$}
& clean & \multicolumn{2}{c|}{$\ell_\infty$}
& clean & \multicolumn{2}{c}{$\ell_\infty$} \\
\cmidrule{3-4} \cmidrule{6-7} \cmidrule{9-10} \cmidrule{12-13}\cmidrule{15-16}
&    &  $\nicefrac{2}{255}$ &  $\nicefrac{4}{255}$
&  & $\nicefrac{2}{255}$ &  $\nicefrac{4}{255}$
&  & $\nicefrac{2}{255}$ &  $\nicefrac{4}{255}$
&  & $\nicefrac{2}{255}$ &  $\nicefrac{4}{255}$
&  & $\nicefrac{2}{255}$ &  $\nicefrac{4}{255}$ \\
\midrule
 % clean              & 93.6 &  2.6 &  0.0 & 74.4 &  0.6 &  0.0 & 71.8 &  7.4 &  0.8 & 69.2 &  5.4 &  1.4 & 77.3 &  4.0 &  0.6 \\
 % $\text{AT (\imnet)}^{\nicefrac{8}{255}}$     & 89.0 &	67.8 &	46.4 &	73.2 &	48.6 &	32.4 &	59.6 &	47.8 &	35.0 &	65.2 &	53.6 &	34.4 &	71.8 &	54.5 &	37.1 \\
% $\text{AT (\imnet)}^{\nicefrac{16}{255}}$   & 81.6 &	62.2 &	47.8 &	59.6 &	46.2 &	35.4 &	50.0 &	46.2 &	33.4 &	56.8 &	47.6 &	37.4 &	62.0 &	50.6 &	38.5\\
% $\text{AT (\cciii)}^{\nicefrac{8}{255}}$ & 87.8 &	66.0 &	47.0 &	71.0 &	51.6 &	33.6 &	57.4 &	49.4 &	37.4 &	65.4 &	50.2 &	34.8 &	70.4 &	54.3 &	38.2 \\
% $\text{AT (\cciii)}^{\nicefrac{16}{255}}$ & 81.4 &	64.6 &	 47.4 &	60.2 &	46.6 &	34.6 &	47.6 &	43.8 &	34.8 &	57.4 &	47.6 &	38.4 &	61.7 &	50.7 &	38.8\\

% 10 steps results
clean              & 93.6 & 2.6 & 0.0 & 74.4 & 0.6 & 0.0 & 71.8 & 7.4 & 0.8 & 69.2 & 5.4 & 1.4 & 77.3 & 4.0 & 0.6 \\
 $\text{AT (\imnet)}^{\nicefrac{8}{255}}$     & 89.6 & 69.0 & 48.8 & 72.4 & 51.6 & 32.8 & 62.0 & 48.8 & 35.8 & 64.8 & 51.2 & 35.6 & 72.2 & 55.2 & 38.3 \\
$\text{AT (\imnet)}^{\nicefrac{16}{255}}$   & 83.4 & 66.6 & 50.0 & 61.2 & 47.6 & 37.4 & 50.0 & 47.2 & 35.8 & 59.4 & 48.8 & 39.2 & 63.5 & 52.6 & 40.6 \\
$\text{AT (\cciii)}^{\nicefrac{8}{255}}$ & 87.6 & 67.0 & 46.8 & 72.4 & 52.0 & 32.2 & 57.4 & 49.2 & 37.8 & 63.6 & 49.6 & 34.4 & 70.3 & 54.5 & 37.8 \\
$\text{AT (\cciii)}^{\nicefrac{16}{255}}$ & 81.0 & 63.2 & 45.8 & 60.2 & 47.6 & 36.8 & 47.6 & 46.0 & 35.0 & 59.0 & 48.2 & 37.0 & 61.95 & 51.3 & 38.7 \\

\bottomrule
\end{tabular}
\label{tab:titok_b_cc3m}
\end{table*}

\subsection{\rev{How do unsupervised attacks change token indices?}}
\label{app:indices}

\rev{
To better understand the effect of unsupervised attacks on both clean and adversarially fine-tuned discrete tokenizers, we can track the number of discrete token indices which change after adding the adversarial perturbations.
In Table~\ref{tab:token_indices} we report the average number changed tokens for the original and our fine-tuned TiTok models, which encode each image in 128 discrete tokens, for unsupervised attacks optimized with 100 steps of APGD on 500 images from ImageNet, at $\epsilon \in \{\nicefrac{2}{255}, \nicefrac{4}{255}\}$.
Moreover, we distinguish between the attacks which are successful against the \fuselip models (the version which uses the corresponding tokenizer) and those which are not.
Interestingly, we observe that for the clean tokenizers even unsuccessful attack lead to very different encoding, where the large majority of token indices are changed.
This suggests that the changing the token indices is not sufficient for effective attacks, and supports our strategy of targeting the embedding vectors instead.
Finally, the adversarially trained models provide stable encoding against the unsupervised attacks, demonstrating the effectiveness of fine-tuning.
}

\subsection{\rev{Analysis of attack objective function}}
\label{app:objective_attacks}

\rev{
To empirically investigate which is the most effective approach to generate adversarial attacks against discrete tokenizers, we compare four different objectives to be optimized by APGD (see Eq.~\eqref{eq:unsupervised-objective}):
\begin{enumerate}
    \item $\norm{h_i(x + \delta) - h_i(x)}_2$ (clean and perturbed before quantization, our default version),
\item $\norm{h_i(x + \delta) - q_i(x)}_2$ (clean before quantization and perturbed after quantization),
\item $\norm{q_i(x + \delta) - h_i(x)}_2$ (clean after quantization and perturbed before quantization),
\item $\norm{q_i(x + \delta) - q_i(x)}_2$ (clean and perturbed after quantization),
\end{enumerate} 
where $h_i$ represents the continuous embedding (before quantization) of token $i$ and $q_i$ its quantized counterpart. We note that the gradient information does not change when using pre- or post-quantization vectors because we employ a straight-through estimator, but the value of the loss, which the APGD optimization algorithm uses to select the strongest attack, does change. 
In Table~\ref{tab:objective_attacks}, we compute the robust accuracy given by 100 steps of APGD with each loss variant at different $\epsilon$ values (Imagenette, \fuselipS and \fuselipB).
Option 1, which uses pre-quantization features for both original and adversarial point, achieves the best results (lower robust accuracy) in nearly all cases, which justifies using it as default objective in our unsupervised attacks both for testing and adversarial training.
}

\begin{table*}[!h]
\centering
\small
\caption{\textbf{UniTok Evaluation under unsupervised attacks.} Robustness evaluation of clean and robust \unitok models against unsupervised attacks, 100 steps at different perturbation strengths.}
\begin{tabular}{@{}lrrrrr@{}}
\toprule
 & clean & $\epsilon=\nicefrac{2}{255}$ & $\epsilon=\nicefrac{4}{255}$ & $\epsilon=\nicefrac{8}{255}$ & $\epsilon=\nicefrac{16}{255}$ \\ \midrule
\textbf{ImageNet-1k} &  &  &  &  &  \\
\midrule
Clean & 67.3 & 0.0 & 0.0 & 0.0 & 0.0 \\
$\text{AT}^{\nicefrac{4}{255}}$ & 66.9 & 66.3 & 62.5 & 17.7 & 0.8 \\
$\text{AT}^{\nicefrac{8}{255}}$ & 58.3 & 61.7 & 61.5 & 42.9 & 3.8 \\
$\text{AT}^{\nicefrac{12}{255}}$ & 50.4 & 53.0 & 53.0 & 49.8 & 11.1 \\
$\text{AT}^{\nicefrac{16}{255}}$ & 42.3 & 44.2 & 44.2 & 43.7 & 19.6\\
\midrule
\textbf{Caltech 101} &  &  &  &  & \\
\midrule
Clean & 85.7 & 6.2 & 2.8 & 1.8 & 1.4 \\
$\text{AT}^{\nicefrac{4}{255}}$ & 81.2 & 82.3 & 76.4 & 27.6 & 4.8 \\
$\text{AT}^{\nicefrac{8}{255}}$ & 77.4 & 78.0 & 78.0 & 57.3 & 11.7 \\
$\text{AT}^{\nicefrac{12}{255}}$ & 72.4 & 71.2 & 71.2 & 66.9 & 27.2 \\
$\text{AT}^{\nicefrac{16}{255}}$ & 65.3 & 66.7 & 66.7 & 65.7 & 39.7 \\
\midrule
\textbf{Imagenette} \\
\midrule
Clean & 99.2&	7.1 &	1.4 &	0.2 &	0.6\\
$\text{AT}^{\nicefrac{4}{255}}$ & 99.2 & 99.0 & 97.8 & 58.7 & 8.5 \\
$\text{AT}^{\nicefrac{8}{255}}$ & 97.8 & 97.4 & 97.4 & 88.1 & 29.8 \\
$\text{AT}^{\nicefrac{12}{255}}$ & 95.6 & 94.2 & 94.2 & 91.9 & 55.8 \\
$\text{AT}^{\nicefrac{16}{255}}$ & 92.7 & 90.9 & 90.9 & 90.7 & 70.4\\
\midrule
\textbf{VQAv2} & &  &  &  &  \\
\midrule
Clean & 73.2 & 43.9 & 38.1 & 33.4 & 30.6 \\
$\text{AT}^{\nicefrac{4}{255}}$ & 67.4 & 67.4 & 66.6 & 48.3 & 38.5 \\
$\text{AT}^{\nicefrac{8}{255}}$ & 62.7 & 62.9 & 62.9 & 57.9 & 39.9 \\
$\text{AT}^{\nicefrac{12}{255}}$ & 61.6 & 61.7 & 61.7 & 60.1 & 43.0 \\
$\text{AT}^{\nicefrac{16}{255}}$ & 57.5 & 57.5 & 57.5 & 57.1 & 51.0 \\
\midrule
\textbf{Ok-VQA} & &  &  &  &  \\
\midrule
Clean & 59.6 & 27.5 & 24.9 & 22.8 & 21.8 \\
$\text{AT}^{\nicefrac{4}{255}}$& 53.0 & 53.0 & 51.8 & 34.0 & 24.0 \\
$\text{AT}^{\nicefrac{8}{255}}$ & 49.2 & 49.1 & 49.1 & 41.9 & 28.1 \\
$\text{AT}^{\nicefrac{12}{255}}$ & 46.7 & 46.5 & 46.5 & 46.8 & 33.4 \\
$\text{AT}^{\nicefrac{16}{255}}$ & 45.1 & 45.5 & 45.5 & 45.5 & 35.6 \\
\midrule
\textbf{GQA }& &  &  &  &   \\
\midrule
Clean & 68.0 & 46.2 & 41.8 & 39.8 & 37.0 \\
$\text{AT}^{\nicefrac{4}{255}}$ & 67.0 & 67.0 & 65.4 & 52.6 & 40.6 \\
$\text{AT}^{\nicefrac{8}{255}}$ & 65.4 & 65.4 & 65.4 & 61.0 & 43.2 \\
$\text{AT}^{\nicefrac{12}{255}}$ & 64.2 & 64.2 & 64.2 & 62.8 & 49.2 \\
$\text{AT}^{\nicefrac{16}{255}}$ & 61.2 & 61.2 & 61.2 & 61.2 & 49.4 \\ \bottomrule
\end{tabular}%

\label{tab:unitok_unsupervised_eval}
\end{table*}

\begin{table}[t]
    \centering \small
    
    \caption{\rev{\textbf{Analysis of the attack objective function.} We compare four objective functions, detailed in App.~\ref{app:objective_attacks}, which use different combinations of pre- and post-quantization embedding vectors, for our unsupervised attacks. We report the robust accuracy on Imagenette obtained optimizing each loss version with 100 steps of APGD at different perturbation radii for \fuselipS and \fuselipB.
    Our default version, indicated as Option 1, achieves the best results (lower robust accuracy) in nearly all cases.
    }}
    \label{tab:objective_attacks}
    \begin{tabular}{l | ccc | ccc}
    \toprule
         \multirow{2}{*}{loss version} &  \multicolumn{3}{c}{\fuselipS} & \multicolumn{3}{c}{\fuselipB}\\
         & $\epsilon=\nicefrac{2}{255}$ & $\epsilon=\nicefrac{4}{255}$ & $\epsilon=\nicefrac{8}{255}$ & $\epsilon=\nicefrac{2}{255}$ & $\epsilon=\nicefrac{4}{255}$ & $\epsilon=\nicefrac{8}{255}$ \\
         \midrule
         Option 1. (default) & 70.7 & \textbf{37.8} & \textbf{16.1} & \textbf{70.5} & \textbf{37.6} & \textbf{20.1} \\
         Option 2. & \textbf{70.6} & 39.7 & 17.0 & 71.1 & 42.1 & 21.0 \\
         Option 3. & 76.3 & 50.5 & 20.7 & 78.6 & 46.9 & 22.0\\
         Option 4. & 76.1 & 50.2 & 21.7 & 77.2 & 49.0 & 22.2 \\
         \bottomrule
    \end{tabular}
\end{table}

% \subsection{\rev{How do unsupervised attacks change token indices?}}
% \label{app:indices}

% \rev{
% To better understand the effect of unsupervised attacks on both clean and adversarially fine-tuned discrete tokenizers, we can track the number of discrete token indices which change after adding the adversarial perturbations.
% In Table~\ref{tab:token_indices} we report the average number changed tokens for the original and our fine-tuned TiTok models, which encode each image in 128 discrete tokens, for unsupervised attacks optimized with 100 steps of APGD on 500 images from ImageNet, at $\epsilon \in \{\nicefrac{2}{255}, \nicefrac{4}{255}\}$.
% Moreover, we distinguish between the attacks which are successful against the \fuselip models (the version which uses the corresponding tokenizer) and those which are not.
% Interestingly, we observe that for the clean tokenizers even unsuccessful attack lead to very different encoding, where the large majority of token indices are changed.
% This suggests that the changing the token indices is not sufficient for effective attacks, and supports our strategy of targeting the embedding vectors instead.
% Finally, the adversarially trained models provide stable encoding against the unsupervised attacks, demonstrating the effectiveness of fine-tuning.
% }

\begin{table}[t]
    \centering \small
    \tabcolsep=4pt
    \caption{\rev{\textbf{How unsupervised attacks change token indices.} 
    We report the average number of token indices that change after unsupervised attacks (100 steps of APGD, 500 ImageNet images).
    }}
    \label{tab:token_indices}
    \begin{tabular}{l | cc | cc | cc | cc}
    \toprule
    & \multicolumn{4}{c|}{\textbf{successful attacks}} & \multicolumn{4}{c}{\textbf{unsuccessful attacks}}\\[2mm]
         \multirow{2}{*}{tokenizer} &
         \multicolumn{2}{c}{\fuselip-S} & \multicolumn{2}{c|}{\fuselip-B} & \multicolumn{2}{c}{\fuselip-S} & \multicolumn{2}{c}{\fuselip-B}
         %\multicolumn{2}{c}{\titok-S} & \multicolumn{2}{c|}{\titok-B} & \multicolumn{2}{c}{\titok-S} & \multicolumn{2}{c}{\titok-B}
         \\
         & $\epsilon=\nicefrac{2}{255}$ & $\epsilon=\nicefrac{4}{255}$ & $\epsilon=\nicefrac{2}{255}$ & $\epsilon=\nicefrac{4}{255}$ & $\epsilon=\nicefrac{2}{255}$ & $\epsilon=\nicefrac{4}{255}$ & $\epsilon=\nicefrac{2}{255}$ & $\epsilon=\nicefrac{4}{255}$ \\
         \midrule
         Clean & 124.5 & 127.4 & 122.8 & 127.0 & 76.8 & 125.4 & 118.3 & 126.0 \\
$\text{AT}^{\nicefrac{4}{255}}$ & 0.0 & 0.0 & 0.0 & 105.0 & 0.0 & 0.0 & 0.0 &  18.1 \\
$\text{AT}^{\nicefrac{8}{255}}$ & 0.0 & 0.0 & 0.0 &  0.0 & 0.0 & 0.0 & 0.0 &  0.0\\
$\text{AT}^{\nicefrac{12}{255}}$ & 0.0 & 0.0 & 0.0 &  0.0 & 0.0 & 0.0 & 0.0 &  0.0\\
$\text{AT}^{\nicefrac{16}{255}}$ &0.0 & 0.0 & 0.0 &  0.0 & 0.0 & 0.0 & 0.0 &  0.0\\
         \bottomrule
    \end{tabular}
\end{table}

% \subsection{\rev{Runtime comparison}}
% \label{app:runtime}

% \rev{
% To clarify the efficiency gains of tokenizer-level adversarial fine-tuning, we directly compare the cost of one training step (computing the adversarial points and updating the model weights) of unsupervised and supervised adversarial training under identical settings, i.e. 10 steps of APGD (note that this setup yield almost identical results to the 50 steps we used for the main results), for TiTok/FuseLIP-S on ImageNet.
% We observe that our unsupervised (tokenizer-only) adversarial training takes 1.17 seconds per sample, while supervised (end-to-end) adversarial training 2.56 seconds per sample.
% This result represents a 2.2x reduction in per-sample training time, despite using the same attack setup.
% The speed-up arises because our method backpropagates only through the tokenizer’s encoder (25.8M parameters), while keeping the codebook and downstream classifier frozen. In contrast, end-to-end AT updates the full model (68M parameters), requiring full backward passes through all components.
% }

\section{Use of AI assistants} 

Some sections of the code in this work were developed with the assistance of an AI coding tool (Copilot), and all such code was carefully reviewed and validated. In addition, parts of the manuscript were refined using writing support tools (Grammarly and ChatGPT-5).

\end{document}